\pdfoutput=1

\documentclass[11pt]{article}

\usepackage[final]{acl}

\usepackage{times}
\usepackage{latexsym,amsmath}
\usepackage{booktabs}
\usepackage{array}
\usepackage{booktabs}
\usepackage{makecell}

\usepackage[T1]{fontenc}

\usepackage[utf8]{inputenc}

\usepackage{microtype}

\usepackage{inconsolata}
\usepackage{algorithm}
\usepackage{ulem}
\usepackage{algpseudocode}
\usepackage{graphicx}
\usepackage{hyperref}

\title{LLM-Guided Lifecycle-Aware Clustering of Multi-Turn Customer Support Conversations}

\author{
 \textbf{Priyaranjan Pattnayak},
 \textbf{Sanchari Chowdhuri},
 \textbf{Amit Agarwal},
 \\
 \textbf{Hitesh Laxmichand Patel}
\\
\\
 \text Oracle America Inc.
\\
 \small{
   \textbf{Correspondence:} \href{mailto:priyaranjan.pattnayak@oracle.com}{priyaranjanpattnayak@gmail.com}
 }
}

\begin{document}
\maketitle
\begin{abstract}
Clustering customer chat data is vital for cloud providers handling multi-service queries. Traditional methods struggle with overlapping concerns and create broad, static clusters that degrade over time. Re-clustering disrupts continuity, making issue tracking difficult. We propose an adaptive system that segments multi-turn chats into service-specific concerns and incrementally refines clusters as new issues arise. Cluster quality is tracked via Davies–Bouldin Index (DBI) and Silhouette Scores, with LLM-based splitting applied only to degraded clusters. Our method improves Silhouette Scores by over 100\% and reduces DBI by 65.6\% compared to baselines, enabling scalable, real-time analytics without full re-clustering.

\end{abstract}

\section{Introduction}

Cloud providers handle high volumes of customer chats that often span multiple service areas e.g., compute, networking, and identity. This complexity makes it hard to accurately categorize and track evolving concerns. Traditional clustering methods treat each chat as a single-topic unit and require full re-runs to update, disrupting consistency and hindering long-term tracking.

Exisiting methods like LDA, K-Means, and HDBSCAN \cite{blei2003lda, macqueen1967kmeans, mcinnes2017hdbscan} enable initial topic discovery but produce static, coarse clusters needing manual refinement. While recent work in intent classification and dialogue tracking \cite{ye2024userintent, gu2022multiparty},  aids real-time understanding, it lacks dynamic organization of granular concerns across domains  \cite{zhu2024dialogue}.

We propose an adaptive framework to cluster user concerns from multi-turn chats. LLMs segment conversations into themes, extract concerns, remove duplicates via contrastive filtering, and classify them into service groups. Appendix \ref{sec:Glossary}  defines key terms used throughout the paper, describing the structure and categorization of customer support conversations in cloud service environments. Within each group, HDBSCAN + UMAP \cite{mcinnes2017hdbscan, van2018umap, mcinnes2018umap} creates interpretable topic clusters, labeled using LLMs \cite{ma2024contrastive,pattnayak9339review}. New concerns are incrementally matched; unmatched ones form new clusters when volume permits.

We track DBI and Silhouette Scores \cite{davies1979dbindex, rousseeuw1987silhouette}. to monitor cluster quality. Degraded clusters are flagged using Z-score and cohesion tests, then refined using LLM-based splitting avoiding full re-clustering and preserving cluster identity for stable, actionable insights.

\paragraph{Our Contributions:}
\begin{itemize}
    \item \textbf{LLM-based segmentation of multi-turn chats:} Breaking down complex chats into coherent themes and distinct concerns, using contrastive filtering to remove duplicates.
    \item \textbf{Service group classification and topic clustering:} Assigning concerns to predefined cloud service groups (F1 > 0.85) and forming interpretable topic clusters.
    \item \textbf{Incremental clustering for emerging concerns:} Dynamically assigning new concerns to existing clusters or forming new ones using LLM-based semantic matching, no full re-clustering needed.
    \item \textbf{Metric-driven adaptive refinement:} We monitor DBI, Silhouette, and cohesion scores to detect drift, refining only affected clusters using LLM-based splitting to maintain stability.
Deployed on 90,000+ chats, the system handles 500+ new concerns daily, enabling real-time issue tracking and trend detection without manual labeling or reprocessing.
\end{itemize}

\section{Related Work}

\subsection{Traditional Clustering in Customer Support}
Methods like LDA \cite{blei2003lda} and K-Means \cite{macqueen1967kmeans} are common for text clustering but struggle with fixed topic counts and long-tailed data, limiting their use for evolving support concerns. HDBSCAN \cite{mcinnes2017hdbscan} improves discovery by auto-selecting cluster counts but requires full re-clustering for updates. Density-based approaches \cite{aggarwal2012text}  have also been tried but lack scalability for real-time support scenarios.

\subsection{Multi-Turn Dialogue Understanding and Intent Classification}
Recent work in dialogue modeling focuses on intent classification and conversational state tracking to improve real-time query understanding \cite{ye2024userintent, patel2025sweeval, gu2022multiparty}. While effective for chatbot resolution, these approaches rely on fixed intent categories \cite{pattnayak2025hybrid} ,and lack the ability to form evolving, structured topic clusters. In contrast, our framework segments multi-turn conversations at the concern level, enabling dynamic clustering beyond static intent labels.

\subsection{Embedding-Based Retrieval}
Retrieval-based clustering methods using models like Sentence-BERT (SBERT) \cite{reimers2019sentencebert, ni2021sentence, thakur2021beir, meghwani2025hardnegativeminingdomainspecific} enhance semantic matching but fail to detect cluster degradation from topic drift. While contrastive learning improves text representations \cite{ma2024contrastive, gao2021simcse, gunel2021supervised}, its application \cite{patel2024llm,pattnayak2024survey} in support clustering is limited. We incorporate contrastive filtering to refine extracted concerns prior to clustering, improving semantic coherence.

\subsection{Incremental and Adaptive Clustering}
Prior work on incremental clustering in streaming data \cite{zhang2018incremental, li2021adaptive, rohit2022lifelong,pattnayak2025label},focuses more on classification than maintaining cluster coherence. Approaches using hierarchical adaptation\cite{moseley2017hierarchical, agarwal2024mvtamperbench} or drift detection \cite{gama2014conceptdrift} often require expensive re-computation. In contrast, our method monitors cluster quality and selectively refines only drifting clusters avoiding full re-clustering.

\cite{bentley2016officevoice} introduced Microsoft’s Office Customer Voice system, which clusters short, single-turn feedback for ad-hoc insights. In contrast, our method handles multi-turn conversations through LLM-guided segmentation and lifecycle-aware clustering, enabling incremental refinement: split, merge, and prune, while preserving cluster identity for longitudinal analysis.

\subsection{Our Approach}
Our work introduces an adaptive clustering framework that (1) segments multi-turn chats into themes, removes redundancy and classifies concerns into service groups, (2) dynamically refines clusters through metric-driven monitoring, and (3) leverages LLMs for semantic matching, new cluster creation, and targeted splitting. Unlike prior approaches, we avoid disruptive full re-clustering by continuously tracking DBI, Silhouette, and Cohesion Scores to maintain stable, scalable, and interpretable clustering for customer service analytics.

\section{Methodology}
\label{sec:methodology}
We propose a dynamic clustering framework for customer concerns in multi-turn chats that adapts without full re-clustering. Unlike traditional methods, it incrementally refines clusters while monitoring quality. Table \ref{tab:comparison} compares our method against traditional clustering approaches.

\begin{figure*}[h!]
    \centering
    \includegraphics[width=\textwidth]{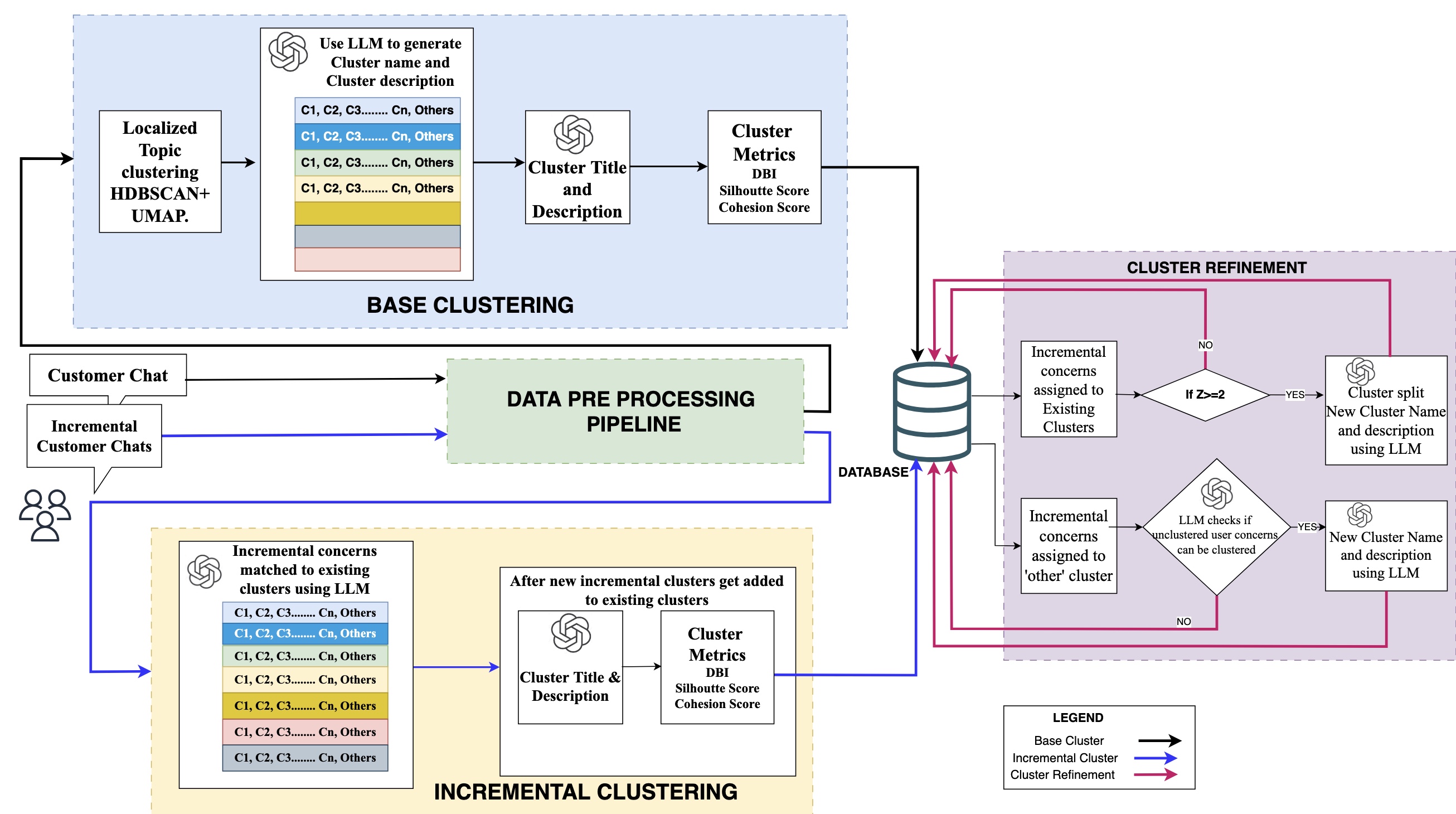}
    \caption{Architecture: The figure outlines base clustering, incremental clustering and cluster refinement pipelines.}
    \label{fig:architecture}
    \vspace{-1em}
\end{figure*}

\begin{table}[h]
\centering
\renewcommand{\arraystretch}{1.2} 
\resizebox{\columnwidth}{!}{ 
\begin{tabular}{p{4.5cm}cc}
\hline
\textbf{Feature} & \textbf{Traditional} & \textbf{Our Approach} \\
\hline
Multi-turn Chat Handling & No & Yes \\
Concern-Level Segmentation & No & Yes \\
Incremental Clustering & Yes & Yes \\
Contrastive Filtering & No & Yes \\
LLM-Based Service Groups & No & Yes \\
Cluster Stability Monitoring & Yes & Yes \\
Automated Cluster Splitting & No & Yes \\
Evolving New Clusters & No & Yes \\
\hline
\end{tabular}
}
\caption{Traditional vs. Proposed Clustering Approach}
\label{tab:comparison}
    \vspace{-1em}
\end{table}
As shown in Figure~\ref{fig:architecture}, the framework has three key phases:(1) base cluster creation via LLM-driven concern extraction, filtering, classification, and HDBSCAN (Phase E of Fig ~\ref{fig:base_workflow}); (2) incremental clustering for new concerns; and (3) LLM-based refinement triggered by cluster drift.

\subsection{{Base Cluster Creation}}

\begin{figure*}[t!]
    \centering
    \includegraphics[width=\textwidth]{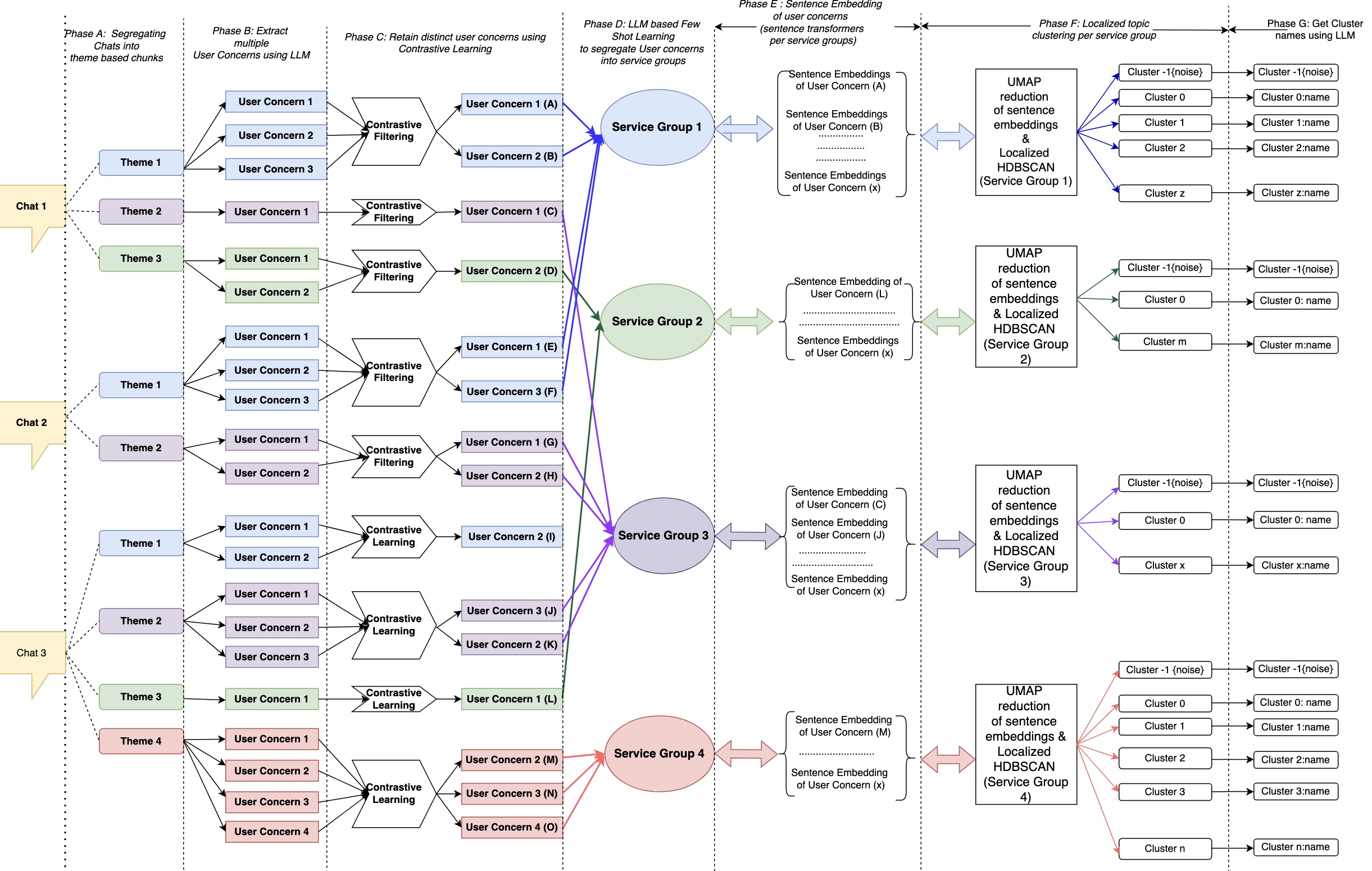}
    \caption{Creation of Base Clusters}
    \vspace{-1em}
    \label{fig:base_workflow}
\end{figure*}

To form initial clusters, we segment multi-turn chats into service-specific themes and extract distinct, contextually grounded concerns. Given the multi-issue nature of chats, this reduces redundancy and preserves clarity. Concerns are then classified into service groups via LLMs and clustered within each group. The full pipeline is shown in Figure.~\ref{fig:base_workflow} where Phase A-E refer the Data Pre processing Pipeline.

\begin{enumerate}
    \item \textbf{Segmentation}: Multi-turn chats are segmented into domain-specific themes using LLM-based detection (Fig.~\ref{fig:base_workflow}, Phase A) with the prompt shown in (Fig.~\ref{fig:LLM Propmt 1} Appendix \ref{sec: appendix LLM propmts}). The model detects topic shifts to separate concerns across service areas. A windowed prompt with overlapping context ensures coherence in long chats. The output is a structured list of segments, each with a theme title and relevant utterances. Quality was validated on 200 chats with strong inter-annotator agreement (Kappa = 0.79), guiding prompt refinement.
    \item \textbf{Concern Extraction}: Each segment is processed by an LLM to extract key user concerns  (Fig.~\ref{fig:base_workflow}, Phase B). A segment may yield multiple granular issues. Since user intents often span multiple utterances, our prompt instructs the model to extract standalone concerns and use windowed context (±1–2 turns) for cross-turn understanding. (e.g., "\textit{My VM crashed and now I can't connect to storage}"), is split into two separate concerns.Sample prompts and output formats are detailed  in (Fig.\ref{fig:LLM Prompt2} Appendix  \ref{sec: appendix LLM propmts}).
We evaluated concern extraction on 150 manually annotated segments labeled by two experts (Kappa = 0.79). The LLM-based method achieved an F1 score of 0.84, indicating strong alignment with human annotations and effective identification of granular user concerns. (Appendix~\ref{sec: appendix_concern_extraction_evaluation}) for performance metrics
    \item \textbf{Contrastive Filtering}:To reduce redundant concerns in clusters, we apply cosine similarity filtering (threshold = 0.95) on \textit{nli-roberta-base-v2} sentence embeddings, as shown in Fig \ref{fig:base_workflow} (Phase C).This step removes duplicate concerns while keeping distinct ones. Since cosine similarity may miss subtle semantic differences, we use a high threshold to minimize false negatives. Within each multi-turn chat, duplicate concerns are removed to prevent overweighting paraphrased repetitions of the same issue (e.g., “login failed… still can’t log in”). The objective of this de-duplication step is to retain only distinct user concerns expressed within a single session, ensuring that intra-chat redundancy does not inflate cluster density. However, identical concerns appearing across different chats are intentionally preserved, as they reflect recurring customer issues and contribute to the representativeness and semantic diversity of the resulting clusters.  (Appendix~\ref{sec:appendix_conrtrastivefiltering}) includes examples of contrastive filtering on semantically similar intents.
    \item \textbf{Service Group Assignment}: Extracted concerns are classified into seven predefined service groups - Compute, Networking, Identity \& Security, Storage, Data Services, Billing \& Accounts, and Others using few-shot LLM prompts (Fig.~\ref{fig:base_workflow}, Phase D); prompt in (Fig.\ref{fig:Prompt3} Appendix  \ref{sec: appendix LLM propmts}).
    \item \textbf{Sentence Embedding and Dimensionality Reduction}: Concerns are embedded using the \texttt{nli-roberta-base-v2} model (768 dimensions) and reduced via UMAP to improve clustering performance and address the curse of dimensionality (Fig.~\ref{fig:base_workflow}, Phases E–F).
    \item \textbf{Localized Clustering}: Concerns within each service group are clustered using HDBSCAN, which identifies topic-based clusters in unsupervised manner as shown in Fig \ref{fig:base_workflow} (Phase F).
    \item \textbf{Cluster Title and Description}: Each cluster is labeled with a title and description using LLM (Fig.~\ref{fig:base_workflow}, Phase G; prompt in (Fig.\ref{fig:Prompt4} Appendix  \ref{sec: appendix LLM propmts}). 
\end{enumerate}
Clustering quality is tracked using DBI, Silhouette Score, and Centroid-Based Cohesion Score.

\subsection{Incremental Clustering}
As new user concerns arise, our framework avoids disruptive full re-clustering by incrementally assigning them to existing clusters. This preserves cluster continuity and reduces computational overhead. To manage evolving concern clusters over time, our framework incorporates a full lifecycle-aware approach including splitting, merging, pruning, role assignment, and drift explanation. The process includes:

\begin{enumerate}
    \item \textbf{Concern Extraction and Filtering}: Incremental concerns are segmented, extracted, contrastively filtered, and classified into service groups using step 1- 4 of Base Cluster creation as described previously. During incremental processing, concern-level de-duplication is applied within each chat to ensure that only unique concerns are forwarded for service group classification and cluster assignment. This step prevents redundant inclusion of paraphrased or repeated issues while preserving cross-session diversity and maintaining a balanced representation of distinct user problems across incremental updates.
    \item \textbf{Cluster Assignment via Hybrid Scoring + LLM Confirmation}: To assign new concerns to the most relevant existing clusters, we employ a hybrid two-stage strategy that balances speed, scalability, and semantic accuracy. Since each incremental concern is already mapped to a specific service group, all matching operations are restricted to clusters within that same group.
\begin{enumerate}
    \item \textbf{Stage 1 – Embedding-Based Similarity Filtering}:Each new concern is encoded using a sentence embedding model (see step 5: Base Cluster creation). Its embedding is compared against precomputed centroid embeddings of cluster using cosine similarity. Top 5 most similar clusters are shortlisted. This step narrows the LLM’s search space, improving efficiency while maintaining high recall.
    \item \textbf{Stage 2 – LLM-Based Semantic Confirmation}:The shortlisted clusters and the new concern are passed into a prompt-driven language model (\textit{cohere.command-r-08-2024 v1.7}). The LLM selects the most appropriate cluster based on natural language understanding, capturing nuance and domain context. It also provides a rationale for its choice to support transparency and auditing.

\end{enumerate}
This \textbf{hybrid scoring + LLM confirmation} approach improves precision for edge cases, emerging topics, and ambiguous inputs, while reducing over-reliance on embeddings or costly full LLM evaluation.
(Fig.\ref{fig:Prompt5} Appendix  \ref{sec: appendix LLM propmts}) .
    \item \textbf{Handling Unassigned Concerns}: Concerns without a match remain in an unclustered pool until enough similar concerns accumulate to form a new cluster, ensuring that emerging topics are identified and tracked over time using LLM. 
\end{enumerate}

LLM-based concern matching offers deeper semantic understanding than embedding-based methods, enabling more accurate and efficient incremental assignments. After incremental assignments are completed, the system recalculates DBI and Silhouette Scores at the service-group level to track overall stability, and updates Centroid-Based Cohesion Scores only for affected clusters. If quality degrades, LLM-driven splitting is triggered to preserve cluster coherence. Fig \ref{fig:Prompt6} in Appendix \ref{sec: appendix LLM propmts} is LLM prompt for splitting a cluster.

\subsection{Evolving New Clusters}

Unassigned concerns are placed in an “Others” pool. After each incremental run, an LLM prompt (Fig.~\ref{fig:Prompt7}, Appendix~\ref{sec: appendix LLM propmts}) checks whether at least 10 similar concerns can form a meaningful new cluster, avoiding low-impact groupings. These new clusters help surface emerging issue types, such as those introduced by new features.

\subsection{Cluster Refinement: Splitting Clusters}

To handle topic drift and evolving concerns, we monitor cluster quality and trigger refinement when needed.A split is initiated when a cluster becomes overly broad, as described in Algorithm~\ref{alg:cluster_split}. A cluster is flagged for review if its service group has DBI > 0.5 or Silhouette < 0.5. We compute the Centroid-Based Cohesion Score:

\begin{equation}
    C_i = \frac{1}{|X_i|} \sum_{x \in X_i} d(x, \mu_i),
\end{equation}
where  
\( C_i \) is the cohesion score for cluster \( i \),  
\( X_i \) is the set of data points in cluster \( i \),  
\( \mu_i \) is the centroid of cluster \( i \), and  
\( d(x, \mu_i) \) is the Euclidean distance between a point \( x \) and the centroid. A high cohesion score indicates dispersed concerns and potential topic drift. We compute a Z-score against historical cohesion to detect abnormal deviations and determine if a split is warranted.
\begin{equation}
    Z_i = \frac{C_i - \mu_{C_{i-1}}}{\sigma_{C_{i-1}}},
\end{equation}
where  
\( C_i \) is the new cohesion score,  
\( \mu_{C_{i-1}} \) is the previous mean cohesion score, and  
\( \sigma_{C_{i-1}} \) is the standard deviation. If \( Z_i \geq 2 \), the cluster is split using an LLM, which reassigns concerns into coherent sub-clusters based on semantic similarity. This preserves explainability while adapting to drift without full re-clustering.

\subsection{Drift Narrative Generation}
To improve explainability, we generate a brief LLM-based narrative after each cluster split. The model is prompted with concerns before and after the split, as well as summaries of the resulting subclusters. It produces a short explanation highlighting the thematic divergence and distinguishing features of the new clusters. These narratives are archived and optionally surfaced in dashboards to help reviewers trace the evolution of concern topics.  Fig \ref{fig:Prompt8- drift narrative} in Appendix \ref{sec: appendix LLM propmts} is LLM prompt for splitting a cluster.

\subsection{Cluster Lifecycle Management: Merge and Prune}
To prevent cluster fragmentation and ensure long-term cohesion, we incorporate a lightweight LLM-guided module to manage cluster lifecycle through merging and pruning. This is especially critical in production settings, where redundant clusters degrade explainability and overwhelm downstream workflows.

Merge candidates are identified using centroid embedding similarity. If two clusters within the same service group have cosine similarity above a threshold (empirically set at 0.92), they are passed to an LLM prompt along with their names, summaries, and sample concerns. The LLM determines whether they are semantically overlapping enough to warrant merging and provides a justification. This process ensures that only truly redundant clusters are consolidated, preserving granularity where needed.

Conversely, clusters that have received no new concern assignments for a 30-day window and fall below a minimum concern count (e.g., 10) are considered for pruning. A secondary LLM check validates whether the cluster represents an outdated or incoherent topic. Pruned clusters are archived but not deleted, maintaining historical traceability.

This merge–prune module maintains a stable, interpretable cluster space over time while minimizing unnecessary fragmentation.  Fig \ref{fig:Prompt9- Cluster Merge} in Appendix \ref{sec: appendix LLM propmts} is LLM prompt for splitting merges.

\subsection{ Cluster Lifecycle Role Assignment}
To enhance explainability and enable long-term cluster lifecycle tracking, we introduce a lightweight cluster role categorization scheme. Each cluster is assigned one of four roles: Core, Emerging, Peripheral, or Deprecated,  based on its age, assignment frequency, semantic cohesion, and drift history. Core clusters represent stable, high-traffic topics with sustained relevance; Emerging clusters are recently formed with rising activity; Peripheral clusters are small or low-cohesion groups; and Deprecated clusters show inactivity or semantic decay. These roles provide downstream users with intuitive life-cycle cues and enable prioritization, monitoring, and dashboard summarization at scale without additional supervision.

\begin{algorithm}
\caption{Triggering a Cluster Split}
\label{alg:cluster_split}
\begin{algorithmic}[1]
\State \textbf{Input:} Service Group \( S \), Cluster \( c \), Cohesion Scores \( C_i \)
\State \textbf{Output:} Updated cluster assignments
\State Get Precomputed DBI and Silhouette Score for \( S \)
\If {DBI \( > 0.5 \) or Silhouette Score \( < 0.5 \)}
    \State Get \(C_i \) for current iteration
    \State Compute \( \mu_{C_{i-1}} \) and \( \sigma_{C_{i-1}} \) for previous iteration from database
    \State Compute Z-score using above values
    \If {\( Z_i \geq 2 \)}
        \State LLM receives title for concerns in \( c \) 
        \State LLM generates new split clusters based on semantic similarity 
        \State Titles and descriptions for split clusters
    \EndIf
\EndIf
\State Return updated clusters
\end{algorithmic}
\end{algorithm}

\section{Experiments and Results}

\subsection{Experimental Setup}

We evaluate our framework on 90,048 anonymized multi-turn chat sessions (Apr-Sep 2024), each tagged into one of seven service groups: Compute, Networking, Identity \& Security, Storage, Billing \& Account, Data Services, and Others. LLM-based segmentation and concern extraction yield almost 148,200 unique concerns for clustering. During Oct-Dec 2024, 400-500 new chats are processed daily via incremental updates.

All LLM tasks (segmentation, concern extraction, service group classification) use the use the \textit{cohere.command-r-08-2024 v1.7} model, selected after comparing four models (\textit{cohere.command-r-plus-08-2024 v1.6, cohere.command-r-08-2024 v1.7, meta.LLaMA 3.3-70B-instruct, meta.LLaMA 3.1-405B-instruct}) on 10,000 chats across service groups Refer Appendix (\ref{sec: LLM_model_evaluation}). To reduce hallucinations, we apply structured prompts, windowed context, contrastive filtering, and post-hoc validation using domain-specific heuristics. Human evaluations further validated LLM reliability (Appendix \ref{sec:appendix_conrtrastivefiltering} and Appendix\ref{sec: appendix_concern_extraction_evaluation}),  ensuring outputs are grounded and cluster-ready. We also track lifecycle transitions, cluster merges, pruning events, and LLM-generated narratives during the 90-day incremental window

\subsection{Evaluation Metrics}

Our pipeline involves both classification and clustering. We use the following standard metrics:
\begin{itemize}
    \item \textbf{Service Group Classification:}
 Precision, Recall, and F1 Score assess the few-shot LLM's ability to assign concerns to correct service groups, ensuring balanced evaluation across categories ~\cite{manning2008introduction}. Service group assignments were further validated using metadata from escalated chats that resulted in formal support tickets, rather than relying on LLM-based judgments. This grounding in verified enterprise outcomes provides an objective benchmark for evaluating classification accuracy and ensures that measured performance reflects real operational correctness.
\item \textbf{Clustering Evaluation:}
 \textbf{Silhouette Score} evaluates intra-cluster similarity vs. inter-cluster difference ~\cite{rousseeuw1987silhouettes}
 \textbf{Davies-Bouldin Index (DBI)} captures cluster compactness and separation (lower is better).
 \textbf{Centroid Based Cohesion Score} tracks internal spread via average distance to centroid, useful for monitoring cluster drift ~\cite{xu2005survey}
 \item \textbf{Lifecycle-Aware Metrics}
 To assess our cluster lifecycle modules such as splitting, merging, pruning, and role tracking, we introduce the following metrics:
\begin{itemize}
    \item Merge Impact ($\Delta$Silhouette, $\Delta$DBI): Measures improvement in clustering quality after LLM-guided merges.
    \item Cluster Stability: Percentage of clusters persisting across multiple time windows.
    \item Role Distribution: Counts of clusters in each role (Core, Emerging, Peripheral, Deprecated) over time.
    \item Role Transitions: Tracks how clusters evolve across roles (e.g., Emerging → Core).
    \item Drift Narrative Clarity (Optional): Human-rated scores (1–5) evaluating the clarity and insightfulness of LLM-generated drift explanations.
\end{itemize}
\end{itemize}

These metrics offer a lightweight yet effective lens into cluster evolution, explainability, and long-term system robustness.

\subsection{Results \& Discussion}
\begin{figure}[h]
  \centering
  \includegraphics[height=5cm, keepaspectratio]{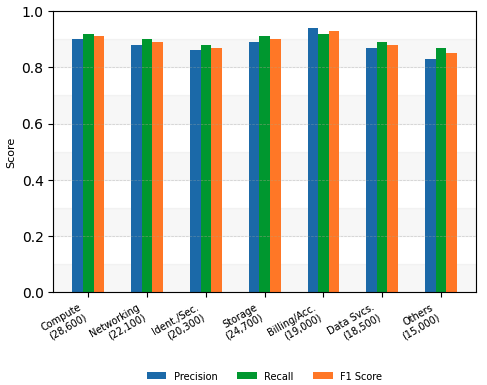}
  \caption{Few-shot classification metrics by Group}
  \label{fig:classification_perf}
  \vspace{-1em}
\end{figure}

Classification Performance and Concern Distribution.
Figure \ref{fig:classification_perf} summarizes few-shot classification metrics across service groups, along with the number of extracted concerns per class.The high F1 scores (>0.85) across all classes indicate the LLM generalizes well, enabling reliable routing to service-group specific clustering.

\textbf{Base Clustering Evaluation}
We apply HDBSCAN with UMAP to perform localized clustering within each service group using sentence embeddings.  Table~\ref{tab:base_clustering} shows strong cohesion and separation across domains.


We apply UMAP + HDBSCAN clustering per service group to retain domain specificity and explainability. Table \ref{tab:base_clustering}, yields strong cohesion and separation improving Silhouette by +0.44 (111.7\%) and reducing DBI by 65.6\% over the global KMeans baseline (Table \ref{tab:baseline_compare}).

\begin{table}[h]
\centering
\resizebox{\columnwidth}{!}{%
\begin{tabular}{lllcc}
\hline
\textbf{\shortstack{Service\\Group}}  & \textbf{\shortstack{User\\Concern\\ Count}} & \textbf{\shortstack{Cluster\\Count}}& \textbf{Silhouette} & \textbf{DBI} \\
\hline
Compute        & 18765& 59& 0.74 & 0.48 \\
Networking     & 22341& 68& 0.72 & 0.36 \\
Ident./Sec.    & 25865& 119& 0.73 & 0.49 \\
Storage        & 26711& 96& 0.71 & 0.43 \\
Billing/Acc.   & 19876& 89& 0.77 & 0.44 \\
Data Svcs.     & 24598& 78& 0.72 & 0.52 \\
Others         & 10044& 105& 0.69 & 0.558 \\
\hline
\textit{Average} & & & \textbf{0.72} & \textbf{0.46} \\
\hline
\end{tabular}
}
\caption{Base clustering metrics - HDBSCAN + UMAP.}
\label{tab:base_clustering}
\vspace{-1em}
\end{table}

\begin{table}[h]
\small
    \centering
    \begin{tabular}{lcc}
    \hline
    \textbf{Method} & \textbf{Silhouette} & \textbf{DBI} \\
    \hline
    KMeans + BERT embeddings & 0.28 & 1.34 \\
    HDBSCAN only & 0.34 & 1.12 \\
    \textbf{HDBSCAN + UMAP} & \textbf{0.72}& \textbf{0.46}\\
    \hline
    \end{tabular}
    \caption{Comparison of clustering methods.}
    \label{tab:baseline_compare}
    \vspace{-1em}
\end{table}

The dataset is moderately balanced across service groups, ranging from 10K (“Others”) to 27K (“Storage”) concerns, forming 59–119 clusters each. This mild imbalance does not affect clustering quality.

HDBSCAN, in combination with UMAP for dimensionality reduction, adapts density-based clustering thresholds locally and does not require uniform cluster sizes. UMAP preserves semantic structure, enabling HDBSCAN to adapt to varying cluster sizes. Average metrics remain stable (Silhouette = 0.72; DBI = 0.46), validating robustness.

UMAP preserves the semantic structure of high-dimensional embeddings, allowing HDBSCAN to discover dense topic-specific groupings even in smaller or sparser groups. As shown in Table~\ref{tab:base_clustering}, clustering performance remains consistent across groups (average Silhouette Score = 0.72; average DBI = 0.46), indicating that our approach is robust to moderate imbalance without requiring  rebalancing.

\begin{figure}[h]
    \centering
    \includegraphics[height=4cm, keepaspectratio]{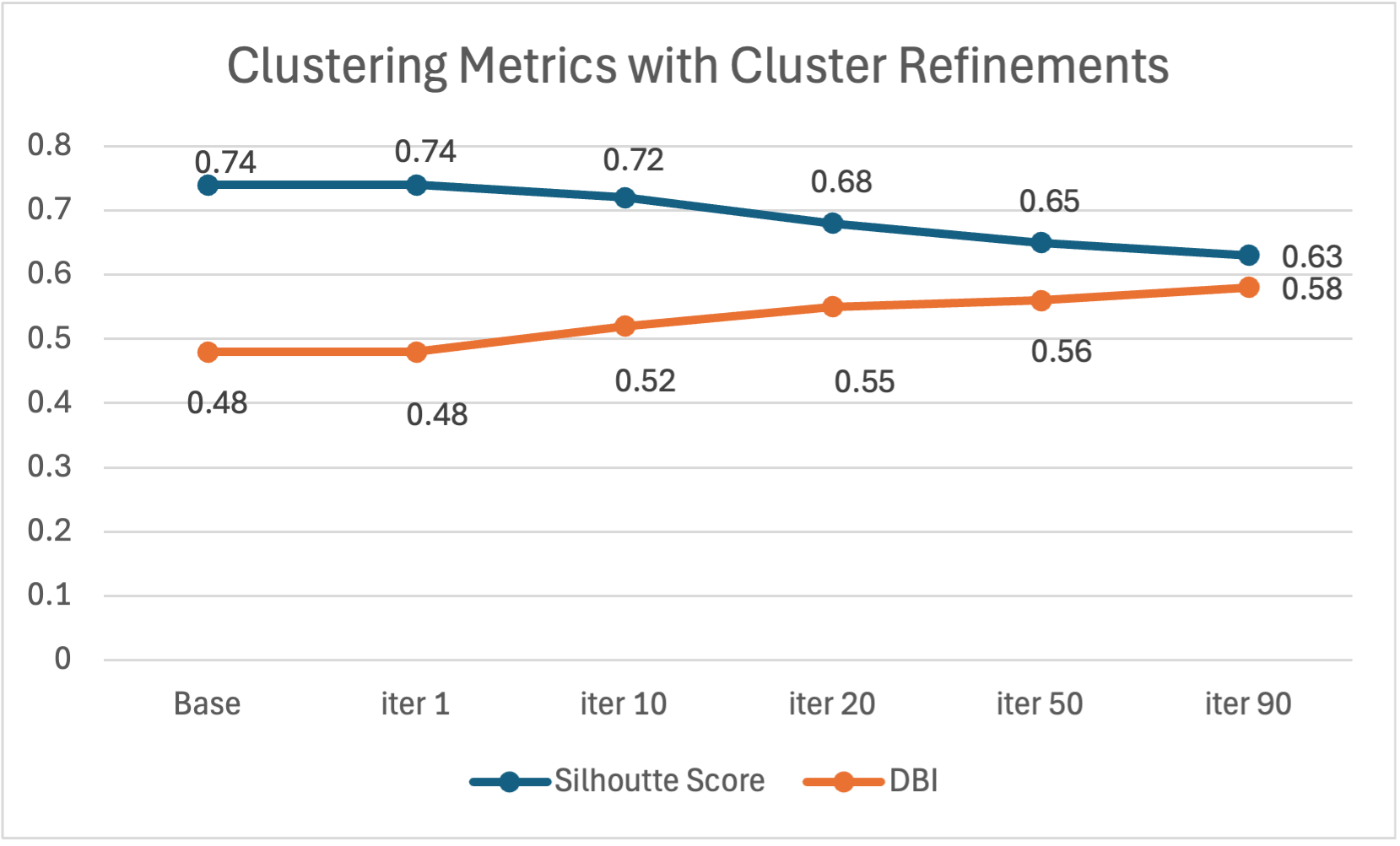}
    \caption{Incremental Clustering Metrics for "Compute with Cluster Refinement" }
    \label{fig:compute_iter}
    \vspace{-1em}
\end{figure}

\begin{figure}
    \centering
    \includegraphics[height=4cm, keepaspectratio]{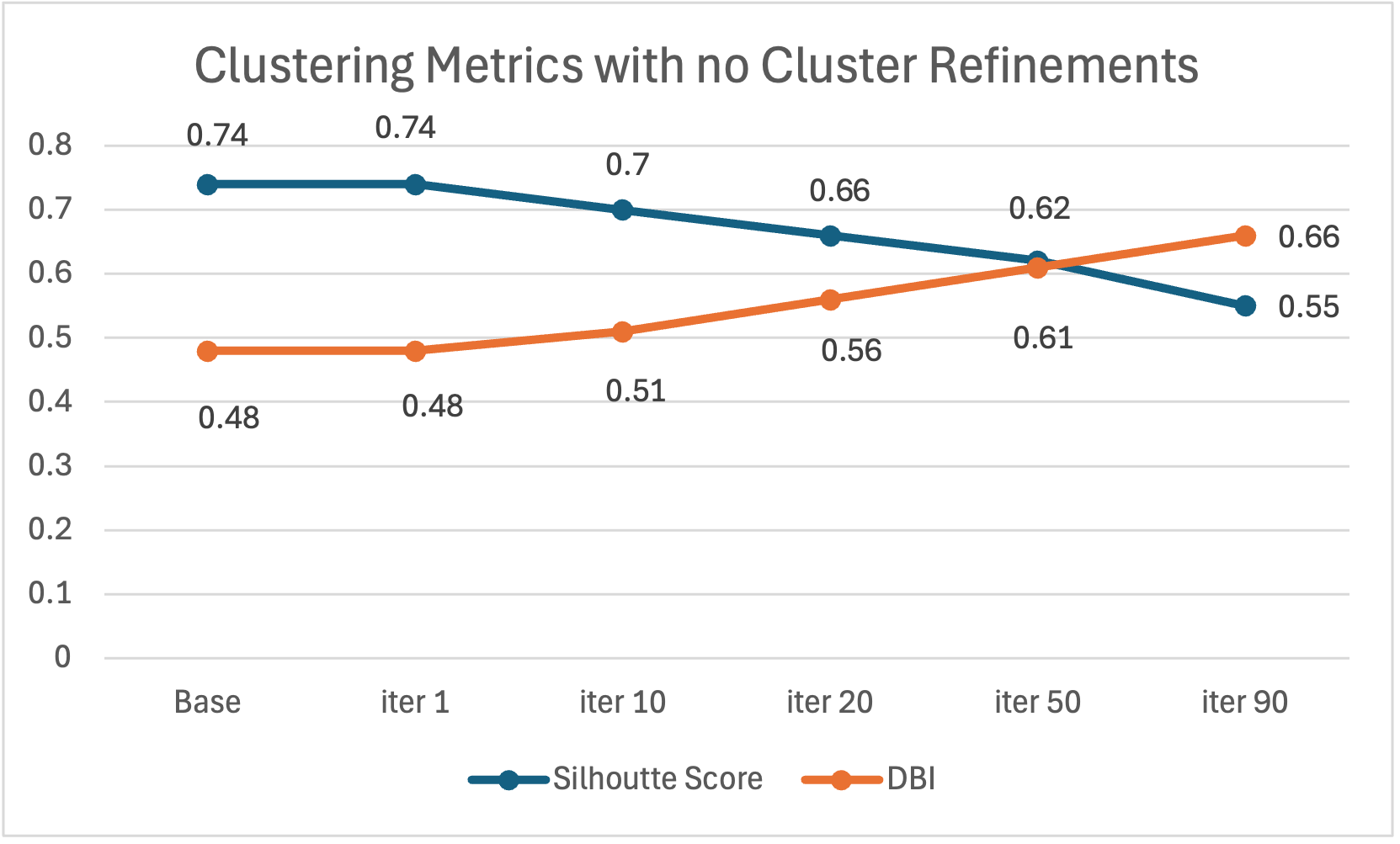}
    \caption{Incremental Clustering Metrics for "Compute" using no Cluster Refinement}
    \label{fig:compute-iter-no-refinement}
\end{figure}

\paragraph{Incremental Clustering Results}
Over 90 days, new chats are incrementally clustered using LLM-based semantic matching without full re-clustering.Figure \ref{fig:compute_iter} shows cluster quality (Compute group) from Day 1 to Day 90. Across service groups, cluster metrics remain stable over time. Appendix \ref{sec:appendix_clustering _metrics_ServiceGroups} Figure \ref{fig:cluster metrics per service group}. Gradual DBI increase and Silhouette decline indicate topic drift; once thresholds are exceeded, LLM-based refinement restores scores within ±20\% of baseline. Figure \ref{fig:compute-iter-no-refinement} confirms quality degrades without refinement, validating the system’s drift detection and adaptive response.


\subsection{Cluster Merge and Prune Evaluation}
\begin{figure}
    \centering
    \includegraphics[height=4cm, keepaspectratio]{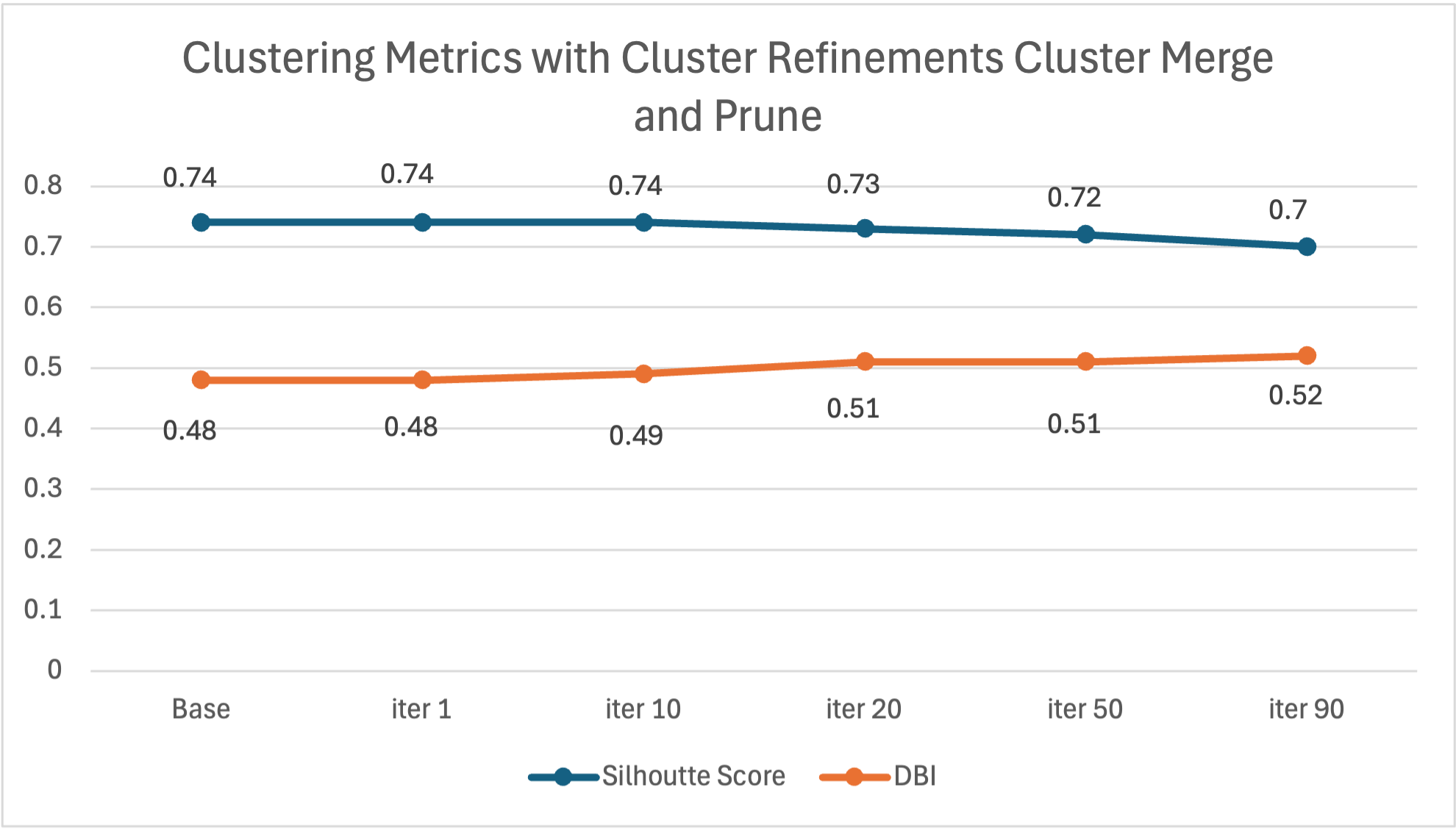}
    \caption{Cluster Merge and Prune}
    \label{fig:Cluster Merge and Prune}
\end{figure}

During the 90-day window, we observed frequent opportunities for consolidation and cleanup. We identified 43 merge candidates using centroid similarity (cos > 0.92), of which 31 were approved by LLM.  Refer Figure \ref{fig:Cluster Merge and Prune} Post-merge, average Silhouette improved from 0.63 to 0.7 and DBI dropped from 0.58 to 0.52 at 90th iteration. This shows that LLM-guided merging removes redundancy and improves cohesion. 12 low-activity clusters were pruned after 30+ days of inactivity and confirmed by LLM to be obsolete.

\subsection{Qualitative Example: Cluster Refinement}

In Billing/Account, a broad cluster covering refunds, invoice errors, and renewals was refined into three distinct sub-clusters. Triggered by cohesion-based Z-score alerts, the LLM-based split improved clarity and downstream labeling. See Appendix~\ref{sec:appendix_qualitative} for full example.

\subsection{Qualitative Example: Cluster Merge and Prune}
In Billing/Account, two clusters one focused on refund delays and the other on non-receipt were identified as semantically overlapping. Despite slight differences in phrasing, both described the same user concern: a refund had been approved but not received. The LLM-guided merge consolidated these into a single, clearer cluster titled “Refund Not Received or Delayed,” reducing redundancy . See Appendix~\ref{sec:appendix_cluster_merge_qualitative} for full example.

\subsection{Topic Drift Frequency and Examples}
Over 90 days, recurring topic drift (Z-score $\geq$ 2) prompted 61 cluster refinements across 7 service groups(5–15 splits each) (avg. 8.7), Data Services and Others has frequent drift. Appendix ~\ref{sec:appendix_splits}

\subsection{Error Analysis and Ablation}
Manual inspection of 600 classification cases revealed most errors (\~60\%) in “Others” and “Data Services,” due to vague or overlapping language. Examples:

\begin{itemize}
    \item \textit{"I’m not getting the output I expected from the portal"} lacks service-specific cues, resulting in misclassification under "Others" instead of "Data Services (Analytics)".
    \item "\textit{I need help with my  pipeline performance}\textbf{"} The phrasing is vague and could refer to  Data Services or Others.
\end{itemize}

Misclassification rate: 7.6\%. Clustering drift was corrected via LLM-based refinement. Ablation studies (Appendix~\ref{sec:appendix_ablation}) show UMAP boosts clustering ($+$0.38 Silhouette, $-$0.66 DBI), and LLM-based matching outperforms cosine similarity (36\% better Silhouette Removing any single component concern extraction, contrastive filtering, or LLM-based classification reduced clustering quality (avg. Silhouette $-$0.21, DBI $+$0.72), confirming each module’s unique value to framework stability and explainability.

\section{Synthetic Dataset Validation}

We have released the synthetic/sanitized dataset\footnote{\href{https://github.com/Synthetic-Datasets-sudo/LLM-Driven-Adaptive-Clustering}{https://github.com/Synthetic-Datasets-sudo}},that is distributionally aligned and replicates the enterprise dataset structure. The synthetic dataset follows the similar service-group distribution as observed in our enterprise corpus (Compute 12.7 \%, Networking 15\%, Identity \& Security 17\%, Storage 18.0 \%, Billing \& Account 15\%, Data Services 16.6\%, Others 6.8\%), yielding approximately 1300-1800 synthetic chats per major service group and 680 for “Others.” This mirrors the 148,000 concern enterprise dataset distribution to maintain clustering comparability.
To evaluate the fidelity of the synthetic dataset, we split the dataset into 8,000 base chats and 2,000 incremental synthetic chats  and compared clustering outcomes of these base chats and incremental synthetic chats against those obtained from the real enterprise dataset. As shown in (Appendix~\ref{sec: Clustering Metrics Synthetic data vs Enterprise Data}) Table \ref{tab:service_metrics}, the synthetic clusters exhibit close alignment in both separation and cohesion metrics (Average DBI = 0.48 (Synthetic dataset)  vs. 0.46 in (Enterprise Data)) and (Average Silhouette = 0.70 (Synthetic dataset) vs. 0.72 (Enterprise Data)).

\begin{figure}[h]
    \centering
    \includegraphics[height=6cm, keepaspectratio]{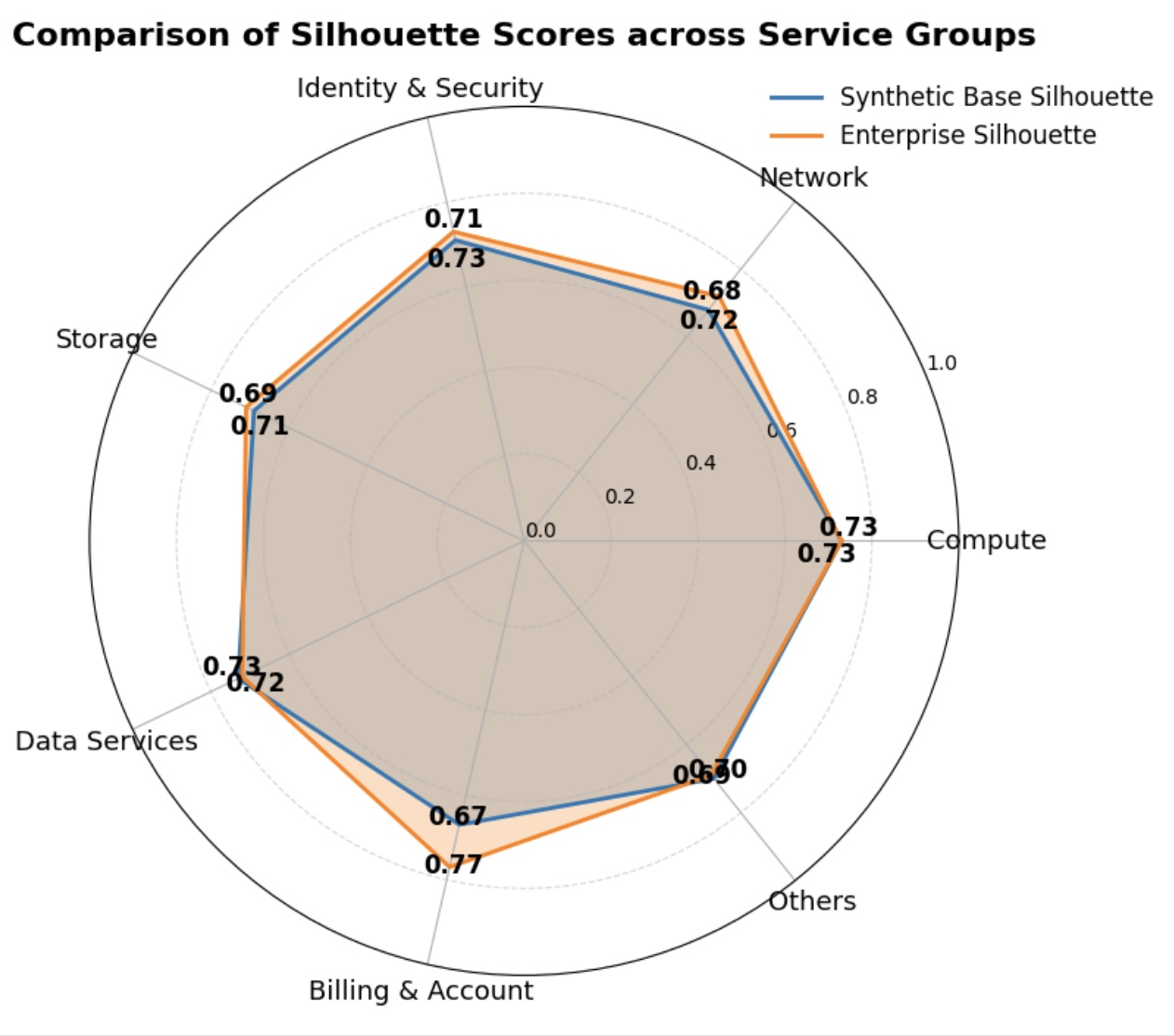}
    \caption{Comparison of Silhouette Scores across Service Groups Enterprise VS Synthetic dataset}
    \label{fig:Radar plot Silhoutte score}
    \vspace{-1em}
\end{figure}

\section{Conclusion}

We present a lifecycle-aware framework for clustering user concerns from multi-turn support chats, combining LLM-based segmentation, contrastive filtering, and unsupervised clustering with adaptive, interpretable refinement. Unlike static or embedding-only methods, our approach supports real-time updates, drift detection, LLM-guided cluster splitting, merging, pruning, and role tracking. Experiments on 90k+ enterprise chats over 90 days demonstrate strong cluster quality, semantic coherence, and robustness to topic drift. Lifecycle operations including 61 splits, 34 merges, and 12 prunes, enable long-term cluster stability while LLM-generated narratives and role labels improve explainability. Together, these components form a scalable, auditable solution for evolving concern management in production environments.

\section*{Limitations}

While the framework shows strong performance, it has a few limitations. Ambiguous or compound concerns spanning multiple services (e.g., compute and storage) can still challenge classification. Evaluation relies primarily on internal metrics and qualitative inspection; future work could include human-in-the-loop or business impact metrics. Finally, the current system supports only English, and extending to multilingual chats is an important next step.

\bibliography{custom}
\appendix
\label{sec:appendix}

\section{Appendix: Terminology and Glossary}
\label{sec:Glossary}
To improve clarity and avoid ambiguity, we define several key terms used throughout the paper. These terms describe the structure and categorization of customer support conversations in cloud service environments.
\begin{itemize}
    \item Multi-turn, Multi-service Chat: A multi-turn, multi-service \cite{pattnayak2025clinicalqa20multitask} chat is a customer support conversation that involves multiple back-and-forth exchanges (multi-turn) between a user and an agent, and spans multiple cloud service areas (multi-service) within the same session.

\textit{For example, a customer might begin a chat about a virtual machine that won’t start (Compute), then ask about related firewall settings (Networking), and finally inquire about unexpected charges (Billing and Account). These topic shifts occur naturally in real-world support chats and pose challenges for traditional clustering methods, which often treat the entire conversation as a single unit.}
\end{itemize}
\begin{itemize}
    \item Theme: A theme is a coherent segment within a multi-turn conversation that centers around a single topic or line of discussion. Themes are extracted by detecting topic shifts, and may contain one or more related concerns. A chat can have multiple 'themes'. In above example, there are three themes: 1) \textit{Compute}, 2) \textit{Networking} and, 3) \textit{Billing and Account}. 
\end{itemize}
\begin{itemize}
    \item Concern: A concern is a distinct user issue, request, or problem described within a customer support conversation. One chat session can contain multiple concerns (e.g., virtual machine boot failure, billing inquiry), each representing a specific topic or task. A theme can have multiple 'concerns'.
\end{itemize}
\begin{itemize}
      \item Service Group (or Domain): A service group (also referred to as a domain) is a predefined cloud service category used to organize concerns (e.g., Compute, Networking, Identity \& Security). Concerns are classified into these groups for structured clustering and analysis \cite{pattnayak2025reviewtoolszerocodellm}.
\end{itemize}

\section{Appendix: Ablation Study of Effect of UMAP}
\label{sec:appendix_ablation}
We see significant improvement, shown in \ref{tab:umap_effect}, in Overall Silhouette Score (+0.38) and DBI improved by 0.66 with UMAP Reduction before clustering as supported by \cite{allaouiumap}. Group wise scores are shown in Table \ref{tab:clustering_scores}.

\begin{table}[h!]
    \centering
    \begin{tabular}{lcc}
    \hline
    \textbf{Configuration} & \textbf{Silhouette} \\[-1ex]
    & \textbf{Score} & \textbf{DBI} \\
    \hline
    HDBSCAN only & 0.34 & 1.12 \\
    HDBSCAN + UMAP & \textbf{0.72} & \textbf{0.46} \\
    \hline
    \end{tabular}
    \caption{Impact of UMAP dimensionality reduction.}
    \label{tab:umap_effect}
\end{table}

\section{Appendix: Ablation Study of LLM Guided Incremental}
\label{sec:appendix_qualitative}
Table \ref{tab:incremental_ablation} shows Clustering metrics of incremental concerns being assigned to Base clusters using LLM based matching (\textit{our proposed solution}) and Centroid based Cosine similarity 
\begin{table}[h]

\centering
\begin{tabular}{lcc}
\hline
\textbf{Method} & \textbf{Silhouette} & \textbf{DBI} \\
\hline
LLM Matching (ours) & 0.72 & 0.46 \\
Cosine Similarity Match & 0.53 & 0.81 \\
\hline
\end{tabular}
\caption{Incremental assignment: LLM-based vs cosine similarity matching on "Compute" Service group on Day 30.}
\label{tab:incremental_ablation}
\end{table}

\section{Appendix: Error Distribution by Service Group}
\label{sec:appendix_error}
We analyzed 600 samples for misclassification and only observed 7.67\% errors, with ~60\% error contributed from Data Services and Others Service Groups as shown in Table \ref{tab:classification_errors}.
\begin{table}[h]
\small
\centering
\begin{tabular}{lcc}
\hline
\textbf{Service Group} & \textbf{Errors (\#)} & \textbf{Error Rate (\%)} \\
\hline
Compute & 4 & 0.67 \\
Networking & 5 & 0.83 \\
Identity \& Security & 3 & 0.50 \\
Storage & 4 & 0.67 \\
Billing/Account & 3 & 0.50 \\
Data Services & 13 & 2.17 \\
Others & 14 & 2.33 \\
\hline
\textbf{600 Samples} & \textbf{46} & \textbf{7.67\%} \\
\hline
\end{tabular}
\caption{Observed classification errors from 600 manually reviewed samples.}
\label{tab:classification_errors}
\end{table}

\section{Appendix: Cluster Split Logs}
We observed a total of 46 cluster splits across all the 7 service groups with highest splits occurring in Data Services which can be attributed to rapidly growing number of services and features in Database and AI services during the last year. Table \ref{tab:split_events} shows group wise splits during the 90 days of incremental clustering.
\label{sec:appendix_splits}

\begin{table}[h]
    \centering
    \begin{tabular}{lc}
    \hline
    \textbf{Service Group} & \textbf{Split Events (90 days)} \\
    \hline
    Compute & 8\\
    Networking & 7 \\
    Identity \& Security & 5 \\
    Storage & 6\\
    Billing \& Account & 9\\
    Data Services & 15\\
    Others & 11\\
    \hline
    \textbf{Total} & \textbf{61}\\
    \hline
    \end{tabular}
    \caption{Number of cluster splits automatically triggered via Z-score.}
    \label{tab:split_events}
\end{table}

\section{Appendix: Cluster Role Categorization and Transitions}
Each cluster is assigned a lifecycle role (Core, Emerging, Peripheral, Deprecated). Table \ref{tab:Cluster Role} shows distribution over time. 68\% of Emerging clusters became Core within 30 days; 14\% of Core clusters transitioned to Deprecated. This confirms the role framework reflects real concern dynamics.
\begin{table}
    \centering
    \begin{tabular}{cccc}\toprule
         Role&  Day 30&  Day 60& Day 90\\\midrule
         Core&  45&  59& 67\\
         Emerging&  22&  16& 11\\
 Deprecated& 3& 8&12\\ \bottomrule
    \end{tabular}
    \caption{Cluster Role Categorization and Transitions}
    \label{tab:Cluster Role}
\end{table}

\section{Appendix: Cluster Refinement Example}
\label{sec:appendix_qualitative}
Table \ref{tab:qualitative_example} shows the case of a cluster within "Billing \& Accounts" service group which had issues from related but different pain points, which resulted in the cluster getting split. Three new clusters were created post refinement process.
\begin{table}[h]
\small
\centering
\begin{tabular}{p{0.45\linewidth} | p{0.45\linewidth}}
\hline
\textit{Before Split} & \textit{After LLM-Driven Split} \\
\hline
Refund delays, invoice errors, auto-renewal issues, discount codes not applying &
New Cluster 1: (Refunds, invoice errors) \newline
New Cluster 2: Auto-renewal failures \newline
New Cluster 3: Discount issues \\
\hline
\end{tabular}
\caption{LLM refinement of a noisy cluster.}
\label{tab:qualitative_example}
\end{table}

\section{{Appendix: Cluster Merge Example}}
\label{sec:appendix_cluster_merge_qualitative}
Table \ref{tab:qualitative example 2} shows two clusters were phrased slightly differently; one emphasizing non-receipt, the other focusing on processing delays. However semantically describe the same user issue: a refund has been requested but hasn’t arrived.

\begin{table}[h]
\small
\centering
\begin{tabular}{p{0.45\linewidth} | p{0.45\linewidth}}
\hline
Before Merge& After LLM- Driven Merge\\
\hline
Base Cluster A: Refund Not Received \newline
Base Cluster B:Delayed Refund Processing&
Refund Not Received or Delayed\\
\hline
\end{tabular}
\caption{LLM driven cluster merge}
\label{tab:qualitative example 2}
\end{table}

\section{Appendix: Base Cluster Creation }
\label{sec:appendix_clustering _metrics_ServiceGroups}
Step by step base cluster creation process is depicted in Fig \ref{fig:base_workflow}. A multi-turn chat is first segregated into theme based chunks in Phase A. In Phase B, an LLM extracts all the user concerns from these themes. There could be several concerns in each theme. Phase C uses contrastive filtering to remove duplicate user concerns from the same theme to ensure we have distinct user concerns per theme. In Phase D, a few-shot LLM classifies the user concerns into one of the 7 different service groups. Phase E generates sentence embedding for the user concerns under each service group followed by Phase F where UMAP reduces the dimensions of embeddings and HDBSCAN clusters user concerns into specific clusters. Finally, in Phase G, LLM generates a cluster title/name and cluster description using user concerns from each cluster.

\begin{table*}[h!]
\centering
\begin{tabular}{lcc}\hline

\textbf{Service Team} & \textbf{Groupwise HDBSCAN} & \textbf{Groupwise HDBSCAN with UMAP} \\\hline

Identity \& Security & 
Silhouette: 0.4984, DBI: 0.7946 & 
Silhouette: 0.7394, DBI: 0.4946 \\

Networking & 
Silhouette: 0.4502, DBI: 0.7452 & 
Silhouette: 0.725, DBI: 0.3652 \\

Billing \& Account & 
Silhouette: 0.4771, DBI: 0.7432 & 
Silhouette: 0.7771, DBI: 0.4432 \\

Storage & 
Silhouette: 0.5026, DBI: 0.757 & 
Silhouette: 0.7126, DBI: 0.4357 \\

Compute & 
Silhouette: 0.4689, DBI: 0.7532 & 
Silhouette: 0.7389, DBI: 0.478 \\

Data Services & 
Silhouette: 0.521, DBI: 0.8043 & 
Silhouette: 0.7218, DBI: 0.528 \\

Others & 
Silhouette: 0.4532, DBI: 0.848 & 
Silhouette: 0.6918, DBI: 0.558 \\ \hline

\end{tabular}
\caption{Clustering performance comparison with and without UMAP dimensionality reduction}
\label{tab:clustering_scores}
\end{table*}

\newpage
\section{Appendix: Clustering Metrics of Different Service groups per iteration }
\label{sec:appendix_clustering _metrics_ServiceGroups}
\begin{figure*}
    \centering
    \includegraphics[width=\textwidth]{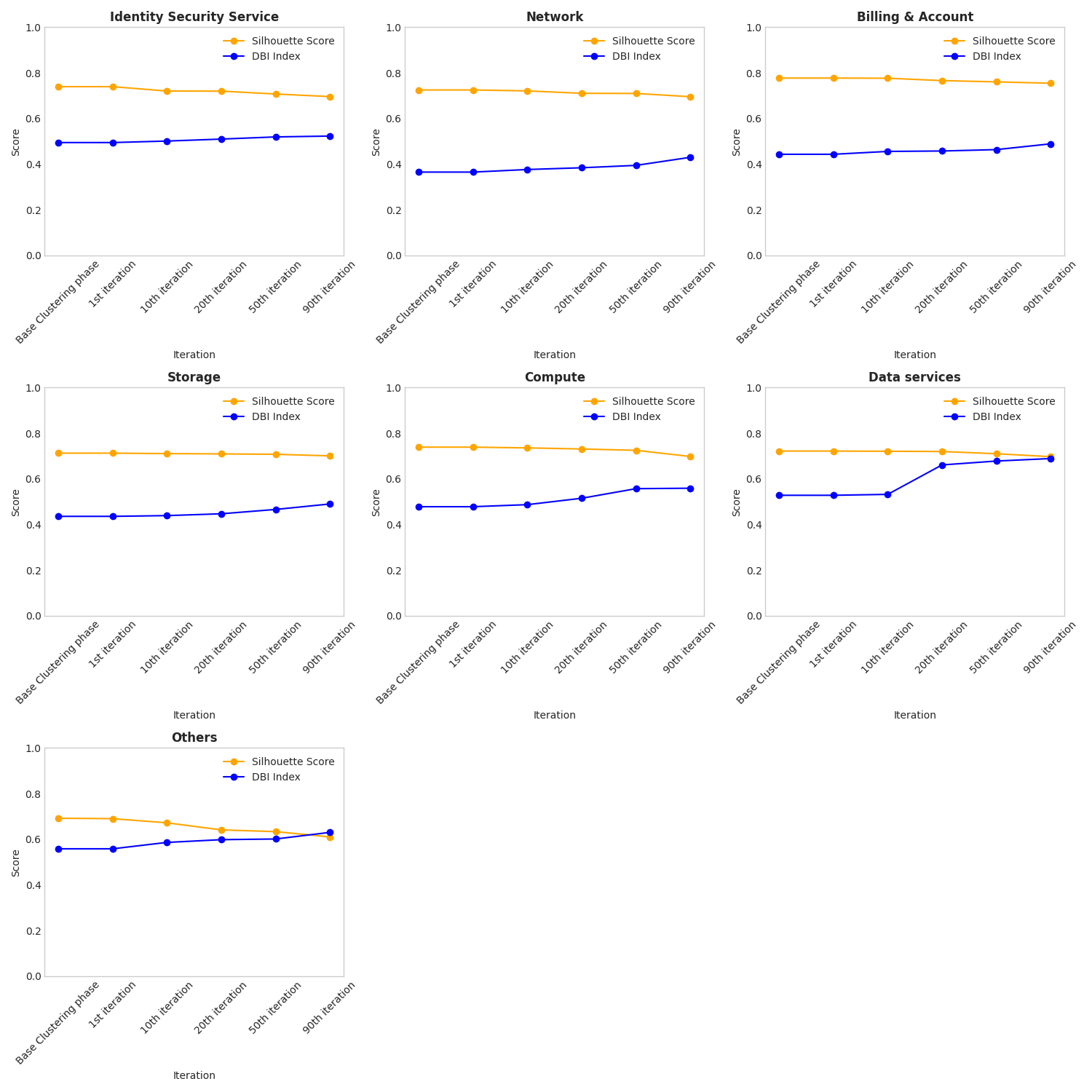}
    \caption{Clustering Metrics of Different Service groups per iteration }
    \label{fig:cluster metrics per service group}
\end{figure*}

We observe that Silhouette Scores and DBI remain stable and within +20\% of initial base cluster values throughout 90 incremental iterations as shown in Fig \ref{fig:cluster metrics per service group}. This proves the efficiency of cluster refinement process which triggers whenever it observes degradation in cluster metrics and quality during incremental step.  

\newpage
\section{Appendix: Examples of Contrastive Filtering on Intents with Similar Semantics}
\label{sec:appendix_conrtrastivefiltering}

The table \ref{tab:contrastive_filtering_examples}  shows how contrastive filtering uses cosine similarity scores to decide which intent pairs to keep. Intents with similarity above 0.95 are considered duplicates, so only one is retained; pairs below this threshold are both kept.
Using an annotated dataset of 100 concerns
with strong inter-annotator agreement (Cohen’s Kappa = 0.79), contrastive filtering correctly identified $89\%$ of semantically similar
intents, effectively removing duplicates and
preserving unique entries.

\begin{table*}[h!]
\centering
\begin{tabular}{p{4cm} p{4.5cm} c c}\hline

\textbf{Reference Intent} & \textbf{Candidate Intent} & \textbf{Cosine Similarity} & \textbf{Action} \\\hline

I want to cancel my subscription. & Please stop my membership. & 0.44 & retain both intents \\
App keeps crashing on startup. & App crashing every time. & 0.956 & retain only 1 intent \\
Change my delivery address. & Update tenancy location. & 0.583 & retain both intents \\
Refund not processed after 7 days. & Still waiting for my refund. & 0.710 & retain both intents \\
Need help with password issues, can't login. & Forgot my password, can't log in. & 0.963 & retain only 1 intent \\
Please reset my account password. & Reset password for account access. & 0.976 & retain only 1 intent \\
\bottomrule
\end{tabular}
\caption{Examples of Contrastive Filtering on Intents with Similar Semantics}
\label{tab:contrastive_filtering_examples}
\end{table*}

\newpage
\section{Appendix: Concern Extraction Evaluation}
\label{sec: appendix_concern_extraction_evaluation}
This appendix presents the evaluation of the LLM-based concern extraction module against a manually annotated dataset of 150 conversation segments. It includes performance metrics and examples comparing model-extracted concerns with ground-truth annotations to illustrate precision, recall, and common error cases.

\begin{table*}[h!]

\centering
\begin{tabular}{p{2.8cm} p{4.5cm} p{4.5cm} c}
\toprule
\textbf{Segment ID} & \textbf{Ground Truth Concerns} & \textbf{LLM-Extracted Concerns} & \textbf{Match Type} \\
\midrule
S001 & 1. VM crash \newline 2. Cannot connect to storage & 1. VM crash \newline 2. Storage access issue & Exact Match \\
S014 & 1. Billing confusion \newline 2. Refund request & 1. Billing confusion \newline 2. Request for compensation & Partial Match \\
S023 & 1. Unable to reset password & 1. Password reset not working & Semantic Match \\
S035 & 1. Data loss after update \newline 2. No backup option & 1. Lost files \newline 2. No backup setting & Exact Match \\
S048 & 1. Login error & – & Missed \\
S057 & – & 1. Account locked & Spurious Output \\
\bottomrule
\end{tabular}
\caption{Concern Extraction Evaluation – Comparison with Annotated Ground Truth}
\label{tab:Concern Extraction Evaluation – Comparison with Annotated Ground Truth}
\end{table*}

\begin{itemize}
    \item \textbf{Exact Match}: Concerns match word-for-word or very closely.
\end{itemize}
    \begin{itemize}
\item \textbf{Semantic Match}: Conceptually equivalent, but phrased differently.
    \end{itemize}
    \begin{itemize}
\item \textbf{Partial Match}: One concern matches, the other is missed or mismatched.
    \end{itemize}
    \begin{itemize}
\item \textbf{Missed}: Model failed to extract any concern present in the ground truth.
    \end{itemize}
    \begin{itemize}
\item \textbf{Spurious Output}: Model generated concern not present in ground truth.

    \end{itemize}

Table \ref{tab:Concern Extraction Evaluation – Comparison with Annotated Ground Truth} compares ground truth concerns with those extracted by the LLM for various conversation segments, categorizing the matches as exact, partial, semantic, missed, or spurious to illustrate extraction accuracy.

 Table\ref{tab:Concern Extraction Evaluation} summarizes the performance of the LLM-based concern extraction module compared to human-annotated ground truth on 150 segments. It reports standard evaluation metrics precision, recall, and F1 score along with 95\% confidence intervals, and shows strong alignment with human annotations. The inter-annotator agreement Cohen's $\kappa = 0.79$ reflects high consistency between annotators, validating the reliability of the dataset.

\begin{table*}[h!]
\centering

    \begin{tabular}{cccc}\toprule
         \textbf{Metric}&  \textbf{Score}&  \textbf{95\% CI}& \textbf{Notes}\\\midrule
         Precision&  0.86&  [0.83, 0.89]& LLM vs. gold-standard concerns\\
         Recall&  0.82&  [0.79, 0.85]& Captures partial \& full matches\\ 
 F1& 0.84& [0.81, 0.86]&Harmonic mean of precision and recall\\
 Inter-Annotator Agreement & 0.79& &Agreement between two human annotators\\\bottomrule
    \end{tabular}
    \caption{Concern Extraction Evaluation against Human-Annotated Data }
    \label{tab:Concern Extraction Evaluation}
\end{table*}

\section{Appendix: Model Selection Study for theme segmentation, concern extraction,  and and service group classification.}
\label{sec: LLM_model_evaluation}

Table \ref{tab:llm_cluster_comparison} shows comparative analysis was conducted across four LLM-based pipelines, each utilizing a different language model to perform the key tasks of \textbf{theme segmentation}, \textbf{concern extraction}, and \textbf{service group assignment}. After these LLM-driven steps, each pipeline applied \textbf{localized clustering} using HDBSCAN with UMAP-based dimensionality reduction to group similar concerns.

To evaluate the clustering quality, we processed 10,000 customer chats spanning 10 service categories through each pipeline. Clustering performance was assessed using two standard metrics: the \textbf{Davies–Bouldin Index (DBI)} and the \textbf{Silhouette Score}. Lower DBI values and higher Silhouette Scores indicate better clustering performance, signifying that the resulting clusters are both compact (internally coherent) and well-separated (distinct from one another).

\begin{table*}[ht]
\centering
\renewcommand{\arraystretch}{1.2}
\begin{tabular}{l c c c c}
\toprule
\textbf{Service Group} & 
\makecell{\textbf{LLaMA 3.3}\\\textbf{(70B)}\\\textbf{DBI / Silhouette}} & 
\makecell{\textbf{LLaMA 3.1}\\\textbf{(405B)}\\\textbf{DBI / Silhouette}} & 
\makecell{\textbf{Cohere R}\\\textbf{DBI / Silhouette}} & 
\makecell{\textbf{Cohere R+}\\\textbf{DBI / Silhouette}} \\
\midrule
Identity \& Security      & 0.44 / 0.71 & 0.43 / 0.73 & \textbf{0.38 / 0.76}& 0.41 / 0.70 \\
Billing \& Account      & \textbf{0.33 / 0.72}& 0.46 / 0.53 & 0.38 / 0.71 & 0.45 / 0.71 \\
Compute                 & 0.57 / 0.69 & 0.35 / 0.53 & \textbf{0.25 / 0.68}& 0.39 / 0.67 \\
Data Services              & 0.72 / 0.38 & 0.68 / 0.52 & 0.62 / 0.50 & \textbf{0.62 / 0.53}\\
Storage                 & \textbf{0.38 / 0.74}& 0.49 / 0.70 & 0.46 / 0.73 & 0.49 / 0.72 \\
Networking         & 0.60 / 0.57 & 0.79 / 0.30 & \textbf{0.45 / 0.68} & 0.45 / 0.65\\
Others                  & 0.56 / 0.65 & \textbf{0.13 / 0.88}& 0.29 / 0.62 & 0.58 / 0.57 \\
\bottomrule
\end{tabular}
\caption{Cluster Quality Comparison across LLM Pipelines (DBI / Silhouette Score). Lower DBI shows clusters are well-separated and internally compact. Higher Silhouette means points are close to their own cluster and far from others.}
\label{tab:llm_cluster_comparison}
\end{table*}

\section{Appendix: Extended Ablation Studies}
This section reports additional ablations evaluating the impact of concern extraction, contrastive filtering, and LLM-based service group classification.
\subsection{Concern Extraction}
Removing LLM-based concern extraction and clustering raw utterances reduces cluster quality, confirming the importance of concern-level segmentation. Table \ref{tab:concern_extraction}  for reference.

\begin{table*}[h!]
\centering
\begin{tabular}{@{}lccc p{1cm}@{}} \toprule
\textbf{Configuration} & \textbf{Silhouette} & \textbf{DBI} & \textbf{Notes}\\
\midrule
Full model (with concerns) & 0.72 & 0.46 & \makecell[l]{HDBSCAN + UMAP baseline\\ with concern extraction}\\
No concern extraction & 0.43 & 1.38 & \makecell[l]{Raw utterances clustered;\\loss of granularity}\\
\bottomrule
\end{tabular}
\caption{Concern Extraction Ablation}
\label{tab:concern_extraction}
\end{table*}

\subsection{Contrastive Filtering}
Omitting duplicate removal degrades cohesion and interpretability. Table \ref{tab:contrastive_filtering}  for reference.

\begin{table*}[h!]
\centering
\begin{tabular}{@{}lccc p{2.8cm}@{}} \toprule
\textbf{Configuration} & \textbf{Silhouette} & \textbf{DBI} & \textbf{Notes}\\
\midrule
Full model (with filtering) & 0.72 & 0.46 & Semantically coherent clusters\\
No filtering & 0.59 & 0.72 & \makecell[l]{Larger, noisier clusters;\\duplicates distort centroids}\\
\bottomrule
\end{tabular}
\caption{Contrastive Filtering Ablation}
\label{tab:contrastive_filtering}
\end{table*}

\subsection{LLM-Based Service Group Classification}
Replacing few-shot LLM classification with embedding-only matching lowers service-group F1 and cluster purity. Table \ref{tab:LLM-Based Service Group Classification}  for reference.

\begin{table*}[h!]
    \centering
    \begin{tabular}{@{}lccc p{2.8cm}@{}} \toprule
         \textbf{Configuration} & \textbf{Silhouette} & \textbf{DBI} & \textbf{\shortstack {Service Group\\ Avg F1}} & \textbf{Notes}\\\midrule
         Full model (LLM classification) & 0.72 & 0.46 & 0.86 & Strong purity, context captured\\
         Embedding-only classification & 0.51 & 1.43 & 0.62 & Misassignments; degraded purity\\ \bottomrule
    \end{tabular}
    \caption{LLM-Based Service Group Classification Ablation}
    \label{tab:LLM-Based Service Group Classification}
\end{table*}

\section {Appendix: Computational Cost and Scalability}

To assess the scalability and practical feasibility of the proposed framework, we report an approximate breakdown of computational cost across three phases: (1) Base Clustering, (2) Incremental Clustering, and (3) Lifecycle Management. All costs are estimated using publicly available token-based \cite{pattnayak2025tokenizationmattersimprovingzeroshot,pattnayak2025tokenization}pricing for comparable LLM APIs and are presented in USD equivalents. Exact pricing cannot be disclosed to preserve vendor confidentiality.

Token Length Considerations: 
\begin{itemize}
    \item Input tokens per call: ~4,500–6,500 (conversation snippet + instructions)
    \item Output tokens per call:~2,000–3,500 (theme/concern/label text)
    \item Total tokens per call: ~6,500–8,500, well within Cohere Command-R’s 16k+ token context window.
\end{itemize}
    
The complete base clustering of approximately 90,000 chats involves around 180,000 LLM API calls, representing a one-time cost of roughly US\$60–\$70. Subsequent incremental clustering operates efficiently, requiring only about 2,250 API calls per day (approximately US\$1.40) to process 500 new chats. Lifecycle management activities—including cluster splitting, merging, pruning, role assignment, and drift narrative generation \cite{agarwal2024enhancing} add roughly 2,400 API calls every 90 days, corresponding to an estimated cost of around US\$1. All prompts and outputs remain well within model context limits, confirming that the overall framework is computationally feasible and scalable for large-scale enterprise deployment.

\subsection{Computation Cost for Base Clustering Phase}
Following are one time costs mentioned in Table \ref{tab:Base Clustering Phase}.It processes ~90k chats.
\begin{table*}[ht]
\centering
\begin{tabular}{lccc}
\hline
\textbf{Step} & \textbf{API Calls} & \textbf{Approx. Tokens / Call} & \textbf{Approx. Cost (US\$)} \\
\hline
Chat $\rightarrow$ Themes & $\sim$90{,}000 & $\sim$6{,}500 input & $\sim$29 \\
Themes $\rightarrow$ Concerns & $\sim$45{,}000 & $\sim$6{,}500 input & $\sim$15 \\
Concern $\rightarrow$ Service Group & $\sim$45{,}000 & $\sim$6{,}500 input & $\sim$15 \\
Cluster Naming \& Description & $\sim$614 & $\sim$2{,}000 output & $\sim$2 \\
\hline
\textbf{Total Base Phase} & --- & --- & \textbf{$\sim$61} \\
\hline
\end{tabular}
\caption{Base Clustering Phase}
\label{tab:Base Clustering Phase}
\end{table*}

\subsection{Computation Cost for Incremental Clustering Phase}
Incremental Clustering phase processes ~500 chats per day. Computation Cost mentioned in Table \ref{tab:Incremental Clustering Phase}.

\begin{table*}[ht]
\centering
\begin{tabular}{lccc}
\hline
\textbf{Step} & \textbf{API Calls} & \textbf{Approx. Tokens / Call} & \textbf{Approx. Cost (US\$)} \\
\hline
Chat $\rightarrow$ Themes & $\sim$500& $\sim$6{,}500 input & $\sim$0.32\\
Themes $\rightarrow$ Concerns & $\sim$250& $\sim$6{,}500 input & $\sim$0.16\\
Concern $\rightarrow$ Service Group & $\sim$250& $\sim$6{,}500 input & $\sim$0.16\\
Cluster Naming \& Description & $\sim$1{,}250& $\sim$6{,}500 input& $\sim$0.80\\
\hline
\textbf{Total Incremental Phase}& --- & --- & \textbf{$\sim$1.44}\\
\hline
\end{tabular}
\caption{Incremental Clustering Phase}
\label{tab:Incremental Clustering Phase}
\end{table*}

\subsection {Lifecycle Management Phase (every 90 days)}
Refer to table \ref{tab:Lifecycle Management Phase}
\begin{table*}[ht]
\centering
\begin{tabular}{lccc}
\hline
\textbf{Step} & \textbf{API Calls} & \textbf{Approx. Tokens / Call} & \textbf{Approx. Cost (US\$)} \\
\hline
Splitting Clusters & $\sim$180 & mix of input + small output ($\sim$6,500 )& $\sim$0.11 \\
Merging Clusters & $\sim$80 & same as $\sim$6,500 & $\sim$0.05 \\
Pruning Clusters & $\sim$12 & small output ($\sim$2,000 tokens)& $\sim$0.01 \\
Role Assignment & $\sim$614 & likely input tokens $\sim$6,500& $\sim$0.20 \\
Drift Narratives & $\sim$120 & small output $\sim$2,000 tokens& $\sim$0.02 \\
\hline
\textbf{\shortstack{Total 90-Day \\Lifecycle Management}} & --- & --- & \textbf{$\sim$0.39} \\
\hline
\end{tabular}
\caption{Lifecycle Management Phase (every 90 days)}
\label{tab:Lifecycle Management Phase}
\end{table*}

\section{Appendix: Clustering Metrics Synthetic data vs Enterprise Data}
\label{sec: Clustering Metrics Synthetic data vs Enterprise Data}
Table \ref{tab:service_metrics}  shows Clustering metrics for Synthetic data and Enterprise Data.

\begin{table*}[h!]
\centering
\renewcommand{\arraystretch}{1.3} 
\large
\begin{tabular}{lcccccc}
\toprule
 & \multicolumn{4}{c}{\textbf{Synthetic Dataset}} & \multicolumn{2}{c}{\textbf{Enterprise Dataset}} \\
\cmidrule(lr){2-5} \cmidrule(lr){6-7}
\textbf{Service} 
 & \textbf{\shortstack{Base\\DBI}} & \textbf{\shortstack{Base\\Silhouette}}& \textbf{\shortstack{Incr\\DBI}} & \textbf{\shortstack{Incr\\Silhouette}}
 & \textbf{\shortstack{Silhouette}} 
 & \textbf{\shortstack{DBI}} \\
\midrule
Compute           & 0.48 & 0.73 & 0.51 & 0.72 & 0.74 & 0.48 \\
Networking        & 0.47 & 0.68 & 0.47 & 0.68 & 0.72 & 0.36 \\
Ident./Sec        & 0.53  & 0.71 & 0.55 & 0.704 & 0.73 & 0.49 \\
Storage           & 0.5 & 0.69 & 0.51 & 0.68 & 0.71 & 0.43 \\
Data Services     & 0.52 & 0.73 & 0.53 & 0.72 & 0.72 & 0.52 \\
Billing account   & 0.43 & 0.67 & 0.52 & 0.65  & 0.77 & 0.44 \\
Others            & 0.44 & 0.7 & 0.45 & 0.71 & 0.69 & 0.558 \\
\bottomrule
\textbf{Average}  & 0.48 & 0.70 & 0.5 & 0.69 & 0.72 & 0.46 \\
\hline
\end{tabular}
\caption{Clustering metrics across service groups for synthetic and enterprise datasets.}
\label{tab:service_metrics}
\end{table*}

\section{Appendix: LLM prompts}
\label{sec: appendix LLM propmts}
\begin{figure*}[t]
    \centering
    \includegraphics[width=1\linewidth]{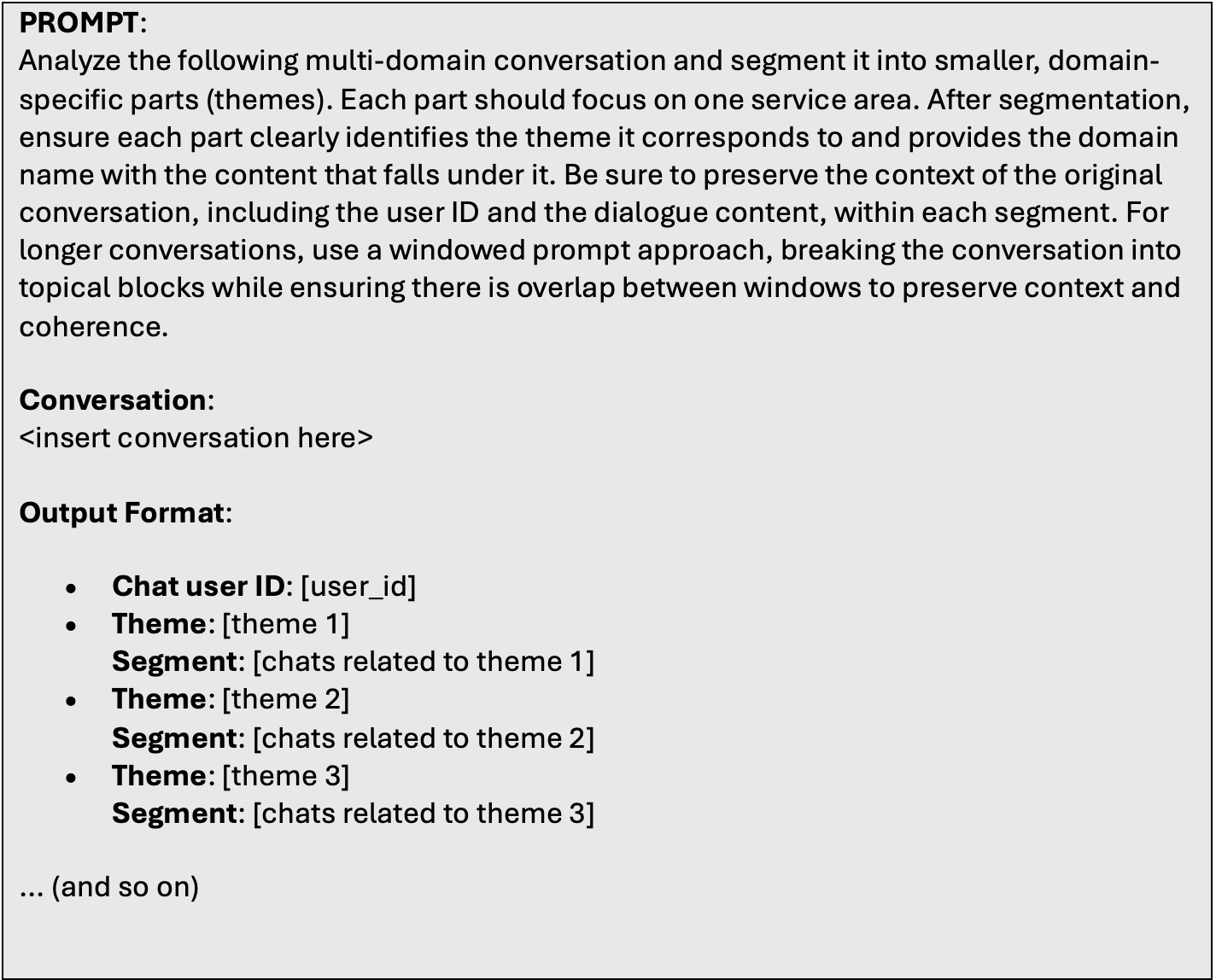}
    \caption{LLM Prompt for segmenting multi service chat into themes}
    \label{fig:LLM Propmt 1}
\end{figure*}

\begin{figure*}[t]
    \centering
    \includegraphics[width=1\linewidth]{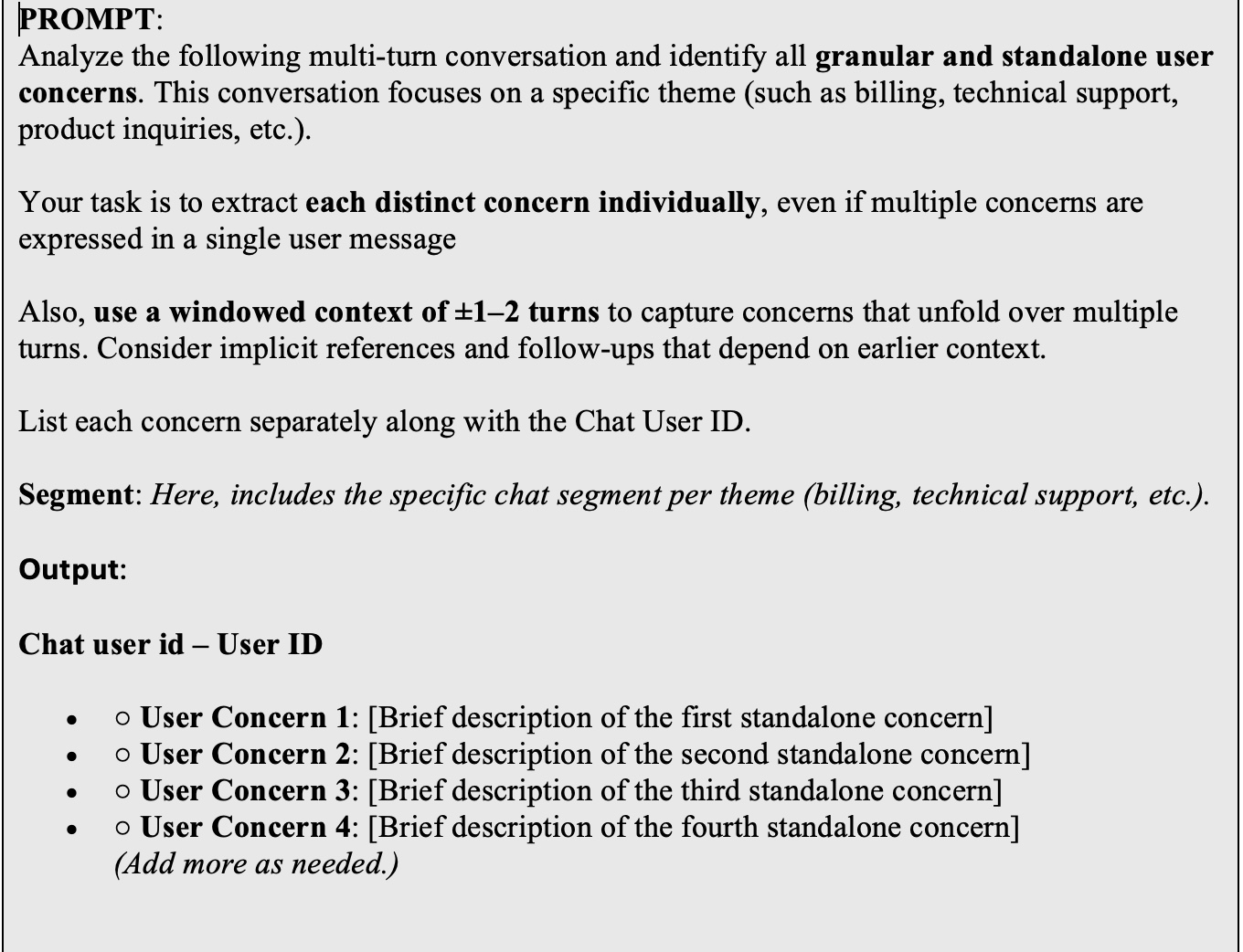}
    \caption{LLM Prompt for extracting user concerns from segmented chats }
    \label{fig:LLM Prompt2}
\end{figure*}

\begin{figure*}[h]
    \centering
    \includegraphics[width=1\linewidth]{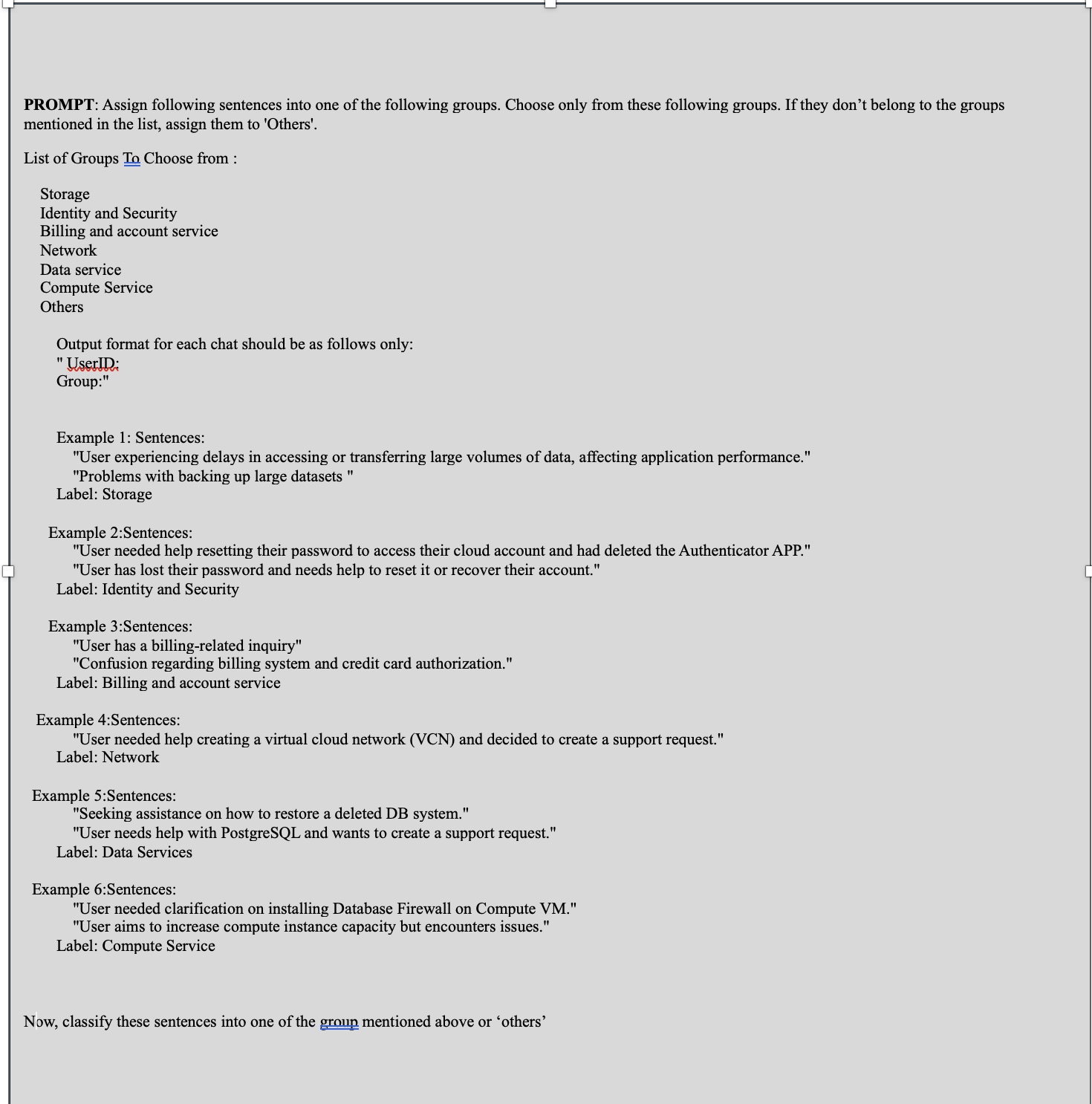}
    \caption{LLM Prompt to assign user concern to service group using LLM based few-shot learning}
    \label{fig:Prompt3}
\end{figure*}

\begin{figure*}[h]
    \includegraphics[width=1\linewidth]{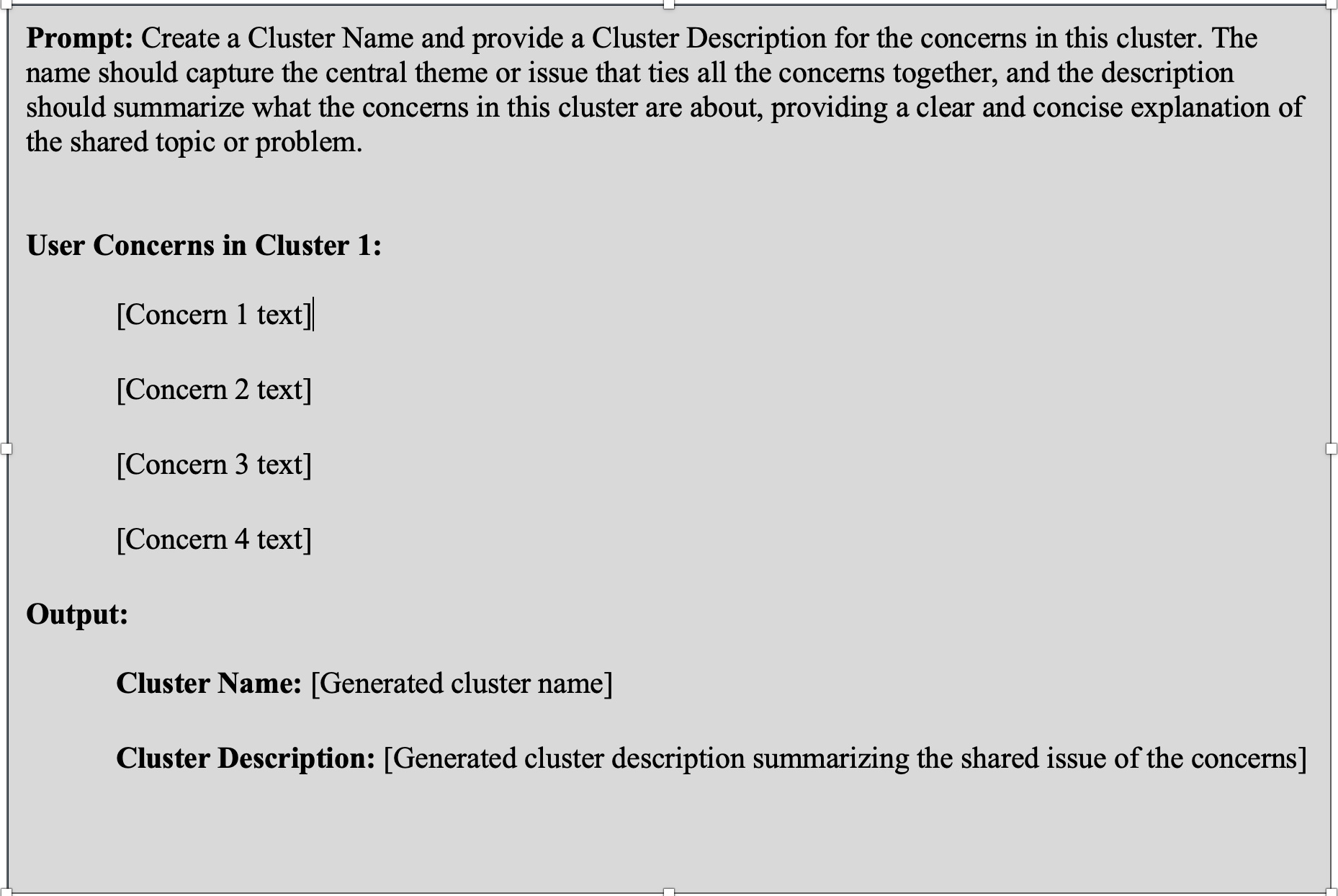}
    \caption{Prompt to Create Cluster Name and Cluster Description}
    \label{fig:Prompt4}
\end{figure*}

\begin{figure*}[h]
    \centering
    \includegraphics[width=1\linewidth]{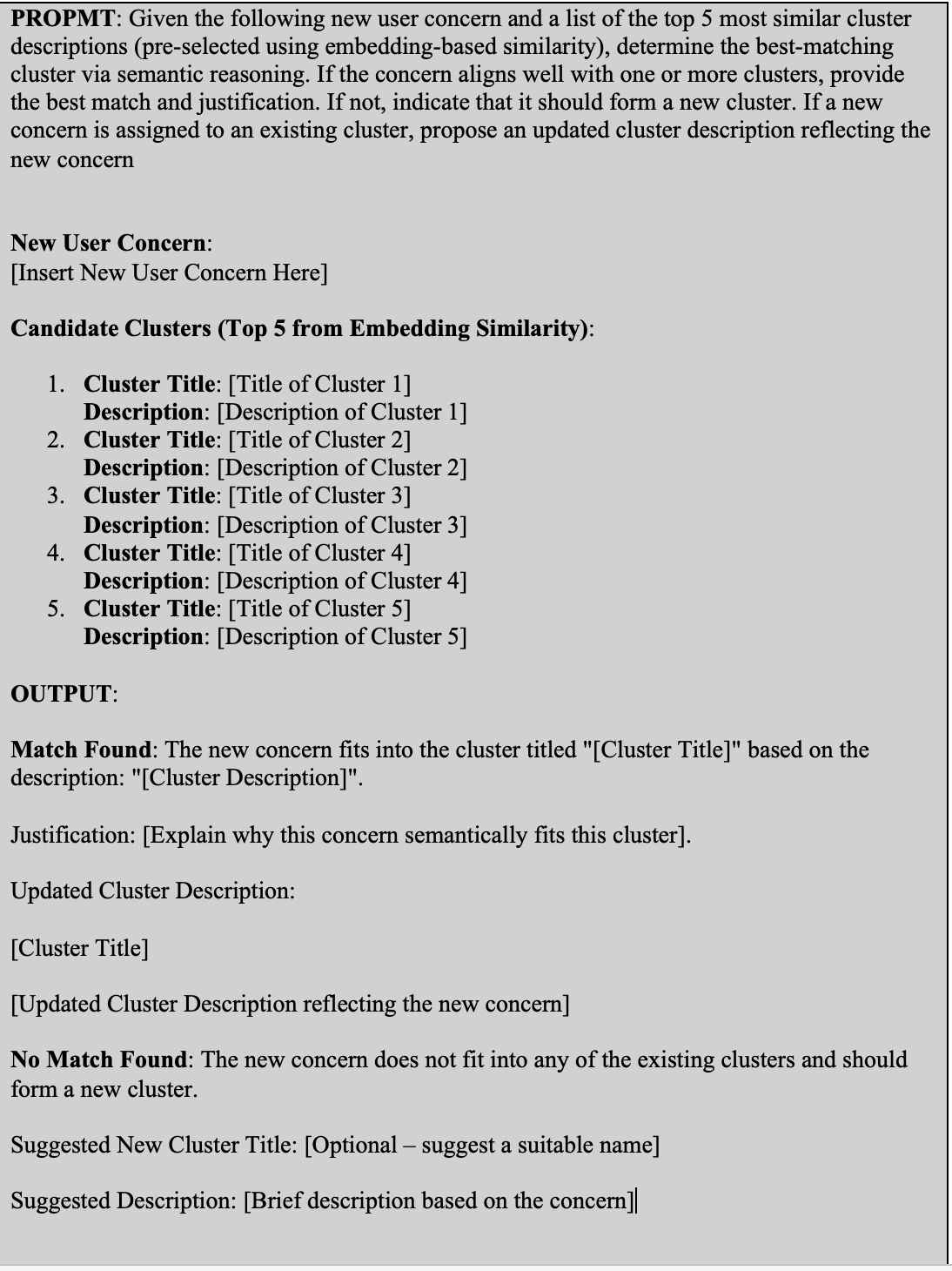}
    \caption{LLM Prompt to assign incremental user concerns to Existing clusters}
    \label{fig:Prompt5}
\end{figure*}

\begin{figure*}[h]
    \centering
    \includegraphics[width=1\linewidth]{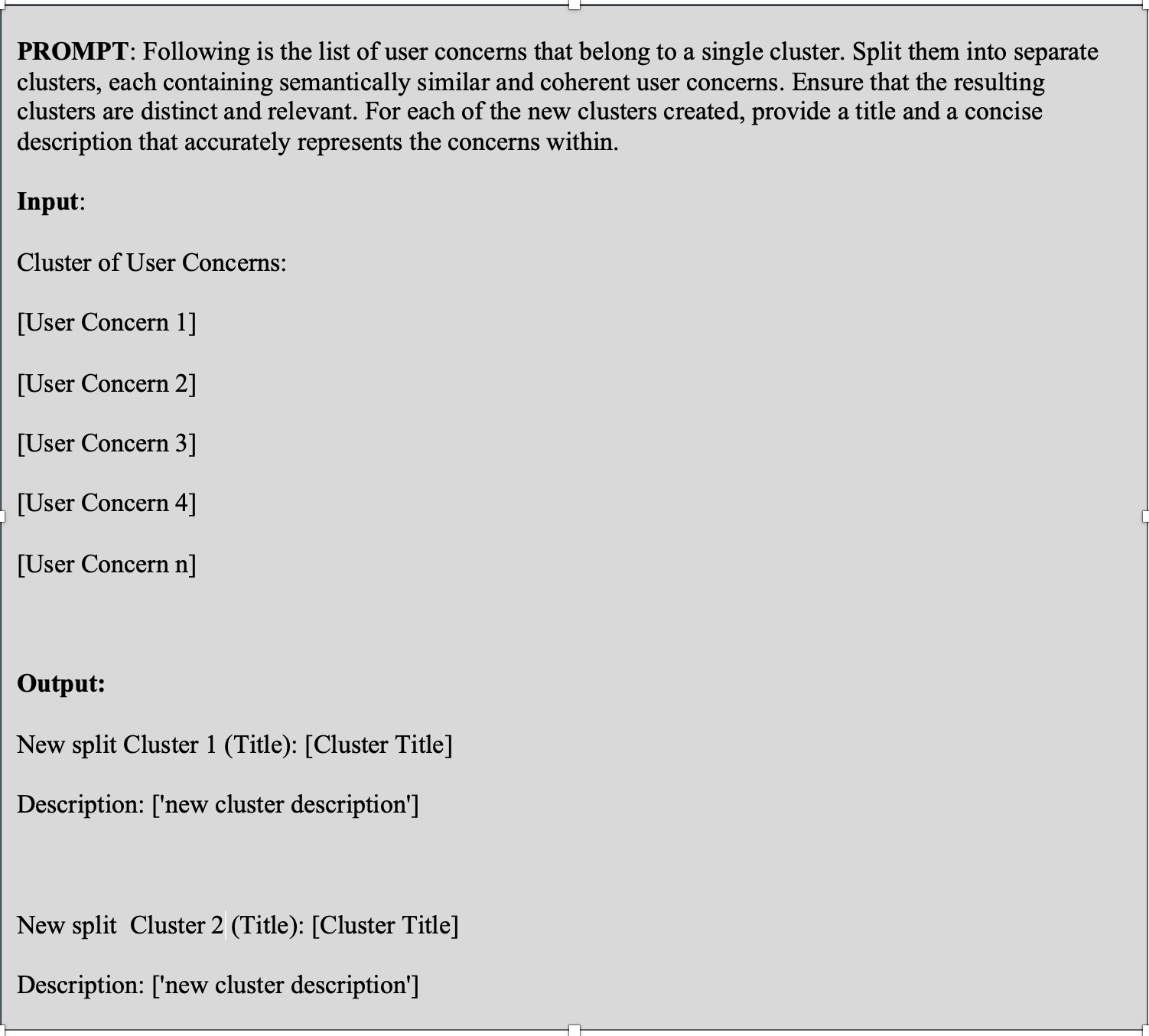}
    \caption{LLM  Prompt for Splitting Cluster}
    \label{fig:Prompt6}
\end{figure*}

\begin{figure*}[h]
    \centering
    \includegraphics[width=1\linewidth]{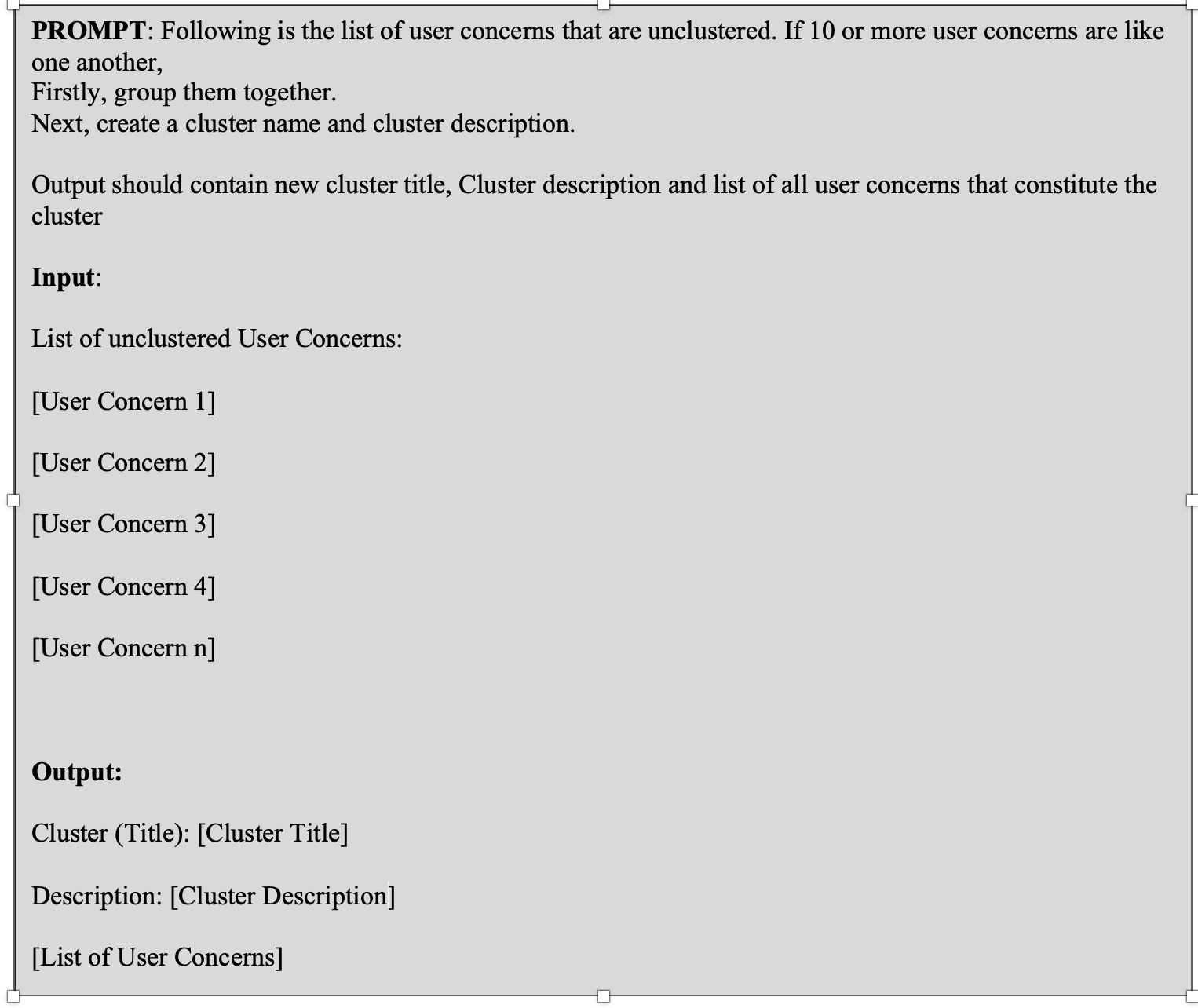}
    \caption{LLM Prompt to Merge Unclustered User Concerns}
    \label{fig:Prompt7}
\end{figure*}

\begin{figure*}[h]
    \centering
    \includegraphics[width=1\linewidth]{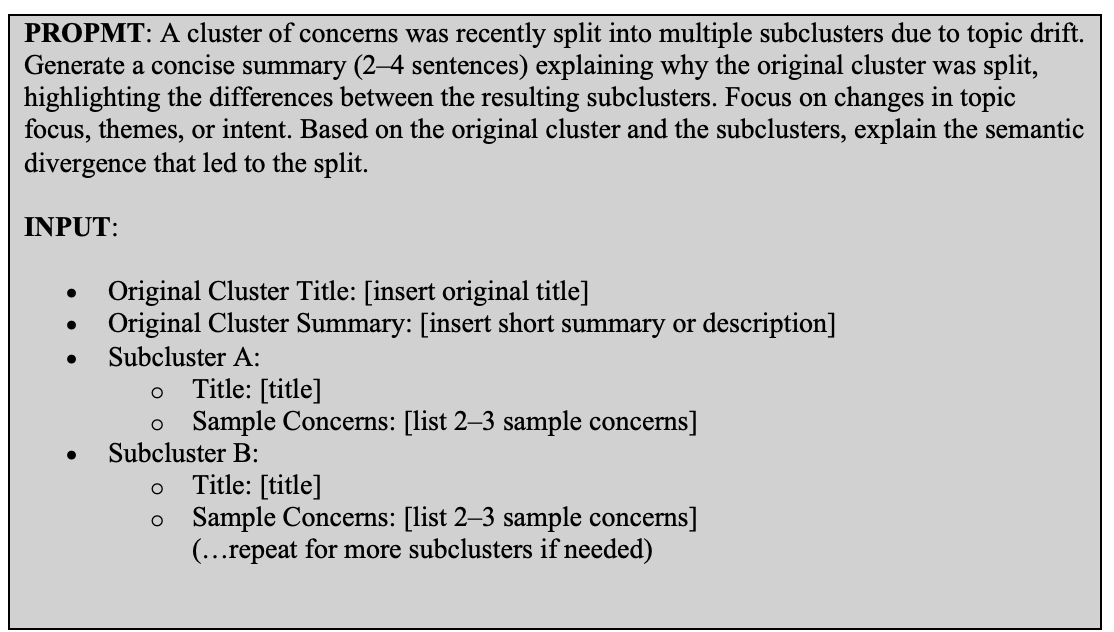}
    \caption{LLM Prompt for Drift Narrative Generation}
    \label{fig:Prompt8- drift narrative}
\end{figure*}

\begin{figure*}[h]
    \centering
    \includegraphics[width=1\linewidth]{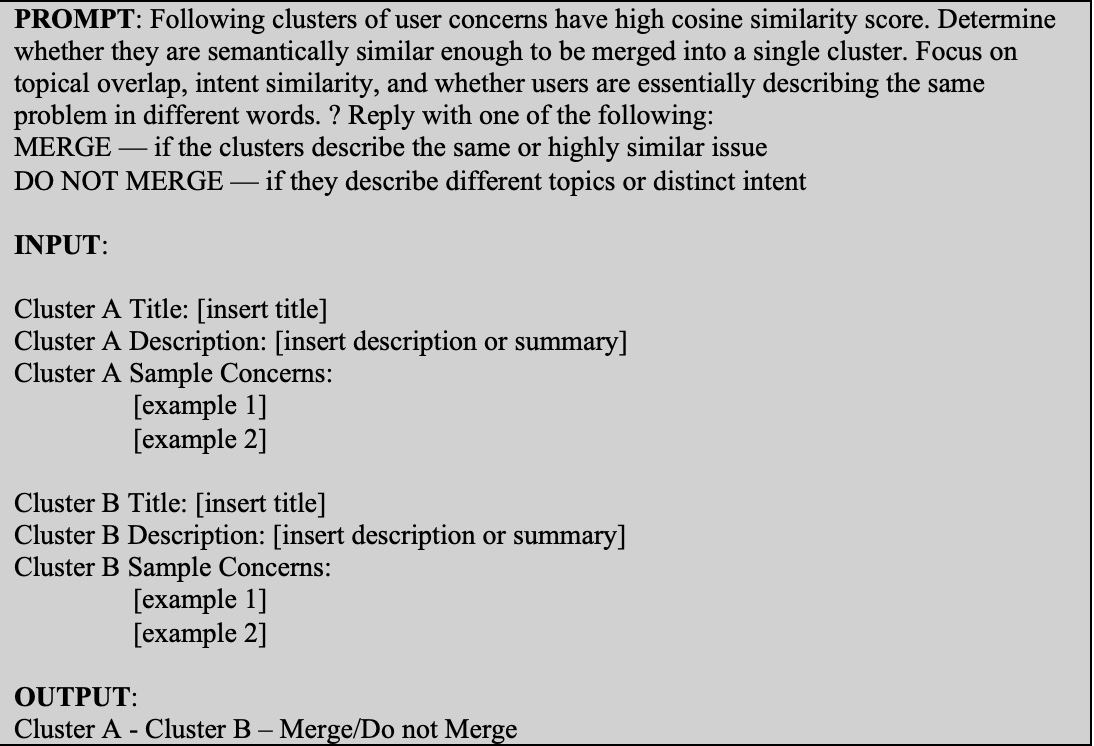}
    \caption{LLM Prompt for Cluster Merge Decision}
    \label{fig:Prompt9- Cluster Merge}
\end{figure*}

\begin{itemize}
    \item Figure\ref{fig:LLM Propmt 1} shows LLM prompt used for extracting themes from multi-service chats
\end{itemize}
\begin{itemize}
    \item Figure \ref{fig:LLM Prompt2} shows LLM prompt used for extracting concerns from segmented chats
\end{itemize}
\begin{itemize}
    \item Figure \ref{fig:Prompt3} shows LLM prompt used for assigning extracted concerns to service groups using LLM based few shot learning
\end{itemize}
\begin{itemize}
    \item Figure \ref{fig:Prompt4} shows LLM prompt used for generating cluster name and cluster description after created using HDBSCAN and UMAP based base clusters.
\end{itemize}
\begin{itemize}
    \item Figure \ref{fig:Prompt5} shows LLM prompt used for assigning incremental concerns to previously created base clusters.
\end{itemize}
\begin{itemize}
    \item Figure \ref{fig:Prompt6} shows LLM prompt used for splitting clusters.
\end{itemize}
\begin{itemize}
    \item Figure \ref{fig:Prompt7} shows LLM prompt used for merging previously unclustered clusters.
    \item Figure \ref{fig:Prompt8- drift narrative} shows LLM prompt used for Drift Narrative Generation
    \item Figure \ref{fig:Prompt9- Cluster Merge} shows LLM prompt used for Cluster Merges
\end{itemize}

\end{document}